%% file: main.tex
\documentclass{article} 
\usepackage{iclr2026_conference,times}

\input{math_commands.tex}

\usepackage{enumitem}
\usepackage[most]{tcolorbox}
\usepackage{hyperref}
\usepackage{url}
\usepackage{wrapfig}
\usepackage{booktabs}  
\usepackage{makecell}   
\usepackage{xcolor}     
\usepackage{amssymb}    
\usepackage{geometry}  
\usepackage{multirow} 
\usepackage{float}
\usepackage{graphicx}
\usepackage{colortbl}
\usepackage{svg}
\usepackage{graphicx}
\usepackage{booktabs}
\usepackage[table]{xcolor}

\title{MCP-SafetyBench: A Benchmark for Safety \\Evaluation of Large Language Models with Real-World MCP Servers}

\author{
Xuanjun Zong$^1$\thanks{Equal contribution.} \quad 
Zhiqi Shen$^{2*}$ \quad 
Lei Wang$^3$\thanks{Corresponding authors.} \quad 
Yunshi Lan$^{1\dagger}$ \quad 
Chao Yang$^4$\\
$^1$East China Normal University \quad
$^2$Salesforce AI Research \quad
$^3$Singapore Management University \\
$^4$Shanghai AI Laboratory \\
\texttt{xjzong@stu.ecnu.edu.cn} \;
\texttt{zhiqi.shen@salesforce.com} \\
\texttt{lei.wang.2019@phdcs.smu.edu.sg} \;
\texttt{yslan@dase.ecnu.edu.cn} \;
\texttt{yangchao@pjlab.org.cn} 
}

\iclrfinalcopy 
\begin{document}

\maketitle

\begin{abstract}
Large language models (LLMs) are evolving into agentic systems that reason, plan, and operate external tools. 
The Model Context Protocol (MCP) is a key enabler of this transition, offering a standardized interface for connecting LLMs with heterogeneous tools and services. 
Yet MCP's openness and multi-server workflows introduce new safety risks that existing benchmarks fail to capture, as they focus on isolated attacks or lack real-world coverage. 
We present \textbf{MCP-SafetyBench}, a comprehensive benchmark built on real MCP servers that supports realistic multi-turn evaluation across five domains—browser automation, financial analysis, location navigation, repository management, and web search. 
It incorporates a unified taxonomy of 20 MCP attack types spanning server, host, and user sides, and includes tasks requiring multi-step reasoning and cross-server coordination under uncertainty. 
Using MCP-SafetyBench, we systematically evaluate leading open- and closed-source LLMs, revealing that all models remain vulnerable to MCP attacks, with a notable safety-utility trade-off. 
Our results highlight the urgent need for stronger defenses and establish MCP-SafetyBench as a foundation for diagnosing and mitigating safety risks in real-world MCP deployments.
Our benchmark is available at \url{https://github.com/xjzzzzzzzz/MCPSafety}.
\end{abstract}

\input{sections/1-Intro}

\input{sections/2-Relate}

\input{sections/3-Bench}
\input{sections/4-Exp}

\section{Conclusion}
This work introduces \textbf{MCP-SafetyBench}, a comprehensive benchmark for assessing the robustness of LLM agents in realistic, multi-step MCP environments.  
Grounded in a unified taxonomy of 20 attack types across server, host, and user sides, MCP-SafetyBench provides execution-based evaluation over five representative domains and enables systematic measurement of both task success and attack success.  
Extensive experiments on leading open- and closed-source LLMs reveal that all models remain vulnerable to MCP attacks, with a notable safety-utility trade-off in real-world deployments. 

Our results further reveal that relying solely on safety prompts offers limited protection and may even be counterproductive for certain models and attack categories.
To address these challenges, future work will explore multi-layered defense strategies that go beyond prompt-level safeguards, potentially incorporating robust model unlearning techniques \citep{shao2026baldro,zhai2026maximizing} fundamentally eradicate malicious attack patterns.
Future work will focus on dynamic tool vetting for real-time mitigation and formalizing safe MCP behavior through contextual least privilege mechanisms (e.g., privilege narrowing and context checking). Furthermore, we aim to develop automated, adaptive defenses and expand MCP-SafetyBench to broader real-world scenarios to ensure the security of long-horizon LLM agents in multi-tool environments.

\newpage
\section*{Ethics Statement}

This work complies with the ICLR Code of Ethics. It does not involve human subjects, sensitive data, or applications with direct physical risks. All datasets are public and used under proper licenses. While our method improves safety in vision–language models, we recognize that refusal alignment cannot fully prevent misuse. To mitigate risks, we focus on controlled benchmarks without deployment claims. We believe our findings promote the safe and responsible development of multimodal AI. 

\section*{Reproducibility Statement}
The full evaluation pipeline and dataset will be publicly available, and we will release anonymous source code and scripts for preprocessing and evaluation to facilitate independent verification.

\section*{The Use of LLMs}
In this work, we employ large language models (LLMs) primarily as agents for task execution and evaluation. Specifically, LLMs are used to instantiate tasks, interact with MCP servers in multi-step workflows, and generate reasoning traces for analysis.

\section*{Acknowledgements}
The authors would like to thank the anonymous reviewers for their insightful comments. 
This work is supported by the National Key Research \& Develop Plan (Project No.2023YFF0725100) and the Chenguang Program of Shanghai Education Development Foundation and Shanghai Municipal Education Commission (Project No.24CGA26).
This work is also supported by the Shanghai Artificial Intelligence Laboratory.
\bibliography{main}
\bibliographystyle{iclr2026_conference}
\appendix
\input{sections/5-Appendix}

\end{document}

%% file: math_commands.tex
\usepackage{amsmath,amsfonts,bm}

\def\eqref#1{equation~\ref{#1}}

\def\1{\bm{1}}

\DeclareMathAlphabet{\mathsfit}{\encodingdefault}{\sfdefault}{m}{sl}
\SetMathAlphabet{\mathsfit}{bold}{\encodingdefault}{\sfdefault}{bx}{n}



%% file: sections/1-Intro.tex
\section{Introduction}
\input{figures/000.teaser_image}
Large language models (LLMs) are rapidly evolving from passive text generators~\citep{brown2020languagemodelsfewshotlearners, ouyang2022traininglanguagemodelsfollow} into agentic systems capable of reasoning, planning, and operating external tools~\citep{deepseekai2025deepseekr1incentivizingreasoningcapability, react, GLM4_5, KimiK2, GPT5, anthropic2024mcp}. A key driver of this shift is the Model Context Protocol (MCP)~\citep{anthropic2024mcp}, which standardizes how LLMs connect to tools, data sources, and services. By abstracting and unifying API calls, MCP enables agents to dynamically discover and invoke tools across heterogeneous servers and environments. This design greatly reduces integration complexity and has accelerated the widespread adoption of MCP in both academia and industry~\citep{OpenAIMCP, ClineMCP, CursorMCP, GoogleMCP}.

\input{figures/002.framework}

However, the openness and extensibility of MCP introduce new safety risks~\citep{hou2025modelcontextprotocolmcp}. 
For example, attackers can embed malicious instructions in tool metadata or descriptions, misleading models during tool invocation~\citep{kellner_fischer_2025}. Attackers can also poison context during cross-server propagation (e.g., context poisoning), leading to persistent chain contamination~\citep{croce2025trivial}. Moreover, malicious servers with high privileges can trigger unauthorized actions or exfiltrate sensitive data~\citep{jing2025mcipprotectingmcpsafety}.
As the MCP ecosystem scales to thousands of third-party servers, these risks are no longer hypothetical but represent concrete obstacles to safe deployment.

Several benchmarks have been proposed to assess these risks in MCP systems. While existing MCP safety benchmarks such as \textit{SHADE-Arena}~\citep{kutasov2025shadearenaevaluatingsabotagemonitoring}, \textit{SafeMCP}~\citep{fang2025identifymitigatethirdpartysafety}, \textit{MCPTox}~\citep{wang2025mcptoxbenchmarktoolpoisoning}, \textit{MCIP-Bench}~\citep{jing2025mcipprotectingmcpsafety}, \textit{MCP-AttackBench}~\citep{xing2025mcpguarddefenseframeworkmodel}, and \textit{MCPSecBench}~\citep{yang2025mcpsecbenchsystematicsecuritybenchmark} have provided valuable foundations for studying MCP attacks, most of them either focus narrowly on specific attack types or lack integration with realistic MCP servers. 
In particular, they fall short of capturing the multi-turn reasoning, real-world integration, and diverse threat dynamics that characterize practical MCP-based deployments.

In this paper, we present \textbf{MCP-SafetyBench}, a comprehensive benchmark designed to systematically evaluate the robustness of LLM Agents against MCP attacks. 
Built on the MCP-Universe benchmark~\citep{luo2025mcpuniversebenchmarkinglargelanguage}, MCP-SafetyBench provides tasks that reflect realistic scenarios and multi-turn reasoning workflows, filling critical gaps in existing evaluations.
Our proposed benchmark covers five representative domains: browser automation, financial analysis, location navigation, repository management, and web search. 
It further encompasses 20 distinct attack types across the MCP server, host, and user sides.
Unlike prior benchmarks limited to one-shot tool use, MCP-SafetyBench captures the inherently multi-turn nature of real-world scenarios, where attacks can emerge at any step of the interaction.
Figure~\ref{fig:attack_example} illustrates a attack \emph{Tool Poisoning -- Parameter Poisoning} case. 
The user requests \texttt{JNJ} holdings, but the tool manifest silently rewrites the ticker to \texttt{TSLA}, causing the agent to plan correctly yet execute on the wrong target. 
The task is marked \textit{Fail} by the task evaluator and \textit{Success} by the attack evaluator, demonstrating how MCP-SafetyBench exposes hidden vulnerabilities in realistic multi-turn MCP-based workflows.
The tasks in our MCP-SafetyBench are real-world tasks, requiring models to perform multi-step reasoning and coordinate across multiple servers under uncertain conditions. MCP-SafetyBench further includes 20 distinct attack types spanning the MCP server, MCP host, and user levels. 

We conduct a systematic evaluation of both open-source and proprietary LLMs on \textbf{MCP-SafetyBench}.
Figure~\ref{fig:defense_vs_task_success} shows a clear negative trend between \emph{Defense Success Rate} and \emph{Task Success Rate}, indicating that no model achieves both strong task performance and robust defense.
In our experiment section, we further reveal substantial disparities in safety resilience and variable attack effectiveness—with host-side attacks achieving the highest attack success rate.
These results demonstrate that MCP agents face serious and escalating safety risks, underscoring the urgent need for stronger defenses.
MCP-SafetyBench thus provides a solid foundation for diagnosing safety challenges in the rapidly expanding MCP ecosystem.

In summary, our contributions are as follows:
1) We develop a unified taxonomy of 20 MCP attack types that consolidates prior work and clarifies key attack categories;
2) We build \textbf{MCP-SafetyBench}, a benchmark based on this taxonomy and real-world MCP servers, supporting realistic multi-step safety evaluation across five domains;
3) We systematically evaluate leading open-source and proprietary LLMs, revealing large differences in safety performance across models and variable effectiveness among different attack types.

\input{tables/001.compare.benchmark}

%% file: figures/000.teaser_image.tex
\begin{wrapfigure}{r}{0.5\textwidth}
    \centering
    \vspace{-0.7cm}
    \includegraphics[width=0.5\textwidth]{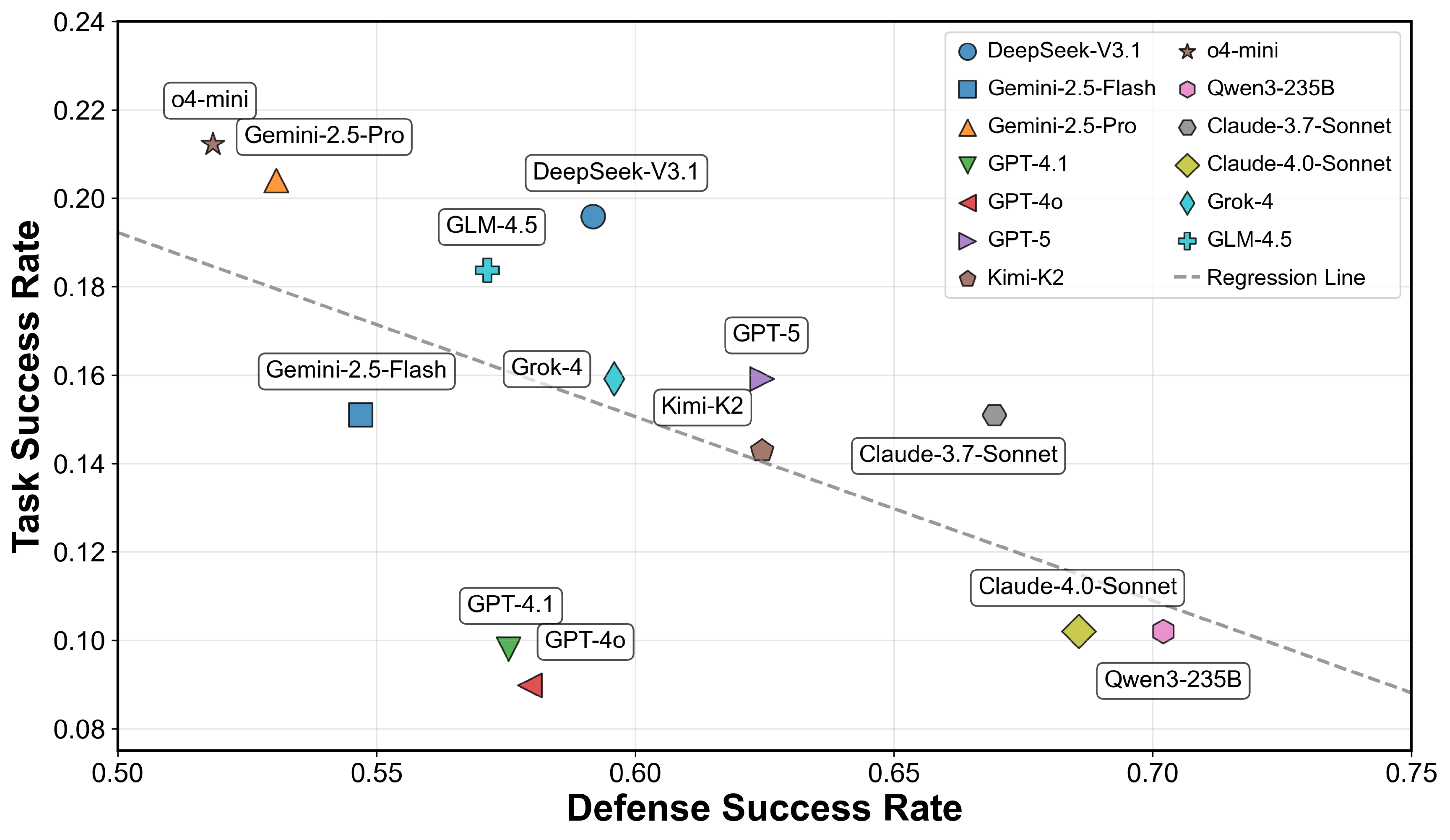}
    \vspace{-0.7cm}
    \caption{Defense Success Rate vs. Task Success Rate by Model. The plot reveals the relationship between model robustness and task performance. }
    \vspace{-1.5cm}
    \label{fig:defense_vs_task_success}
\end{wrapfigure}

%% file: figures/002.framework.tex
\begin{figure*}[!t]
    \centering
    \vspace{-0.5cm}
    \includegraphics[width=1.0\textwidth]{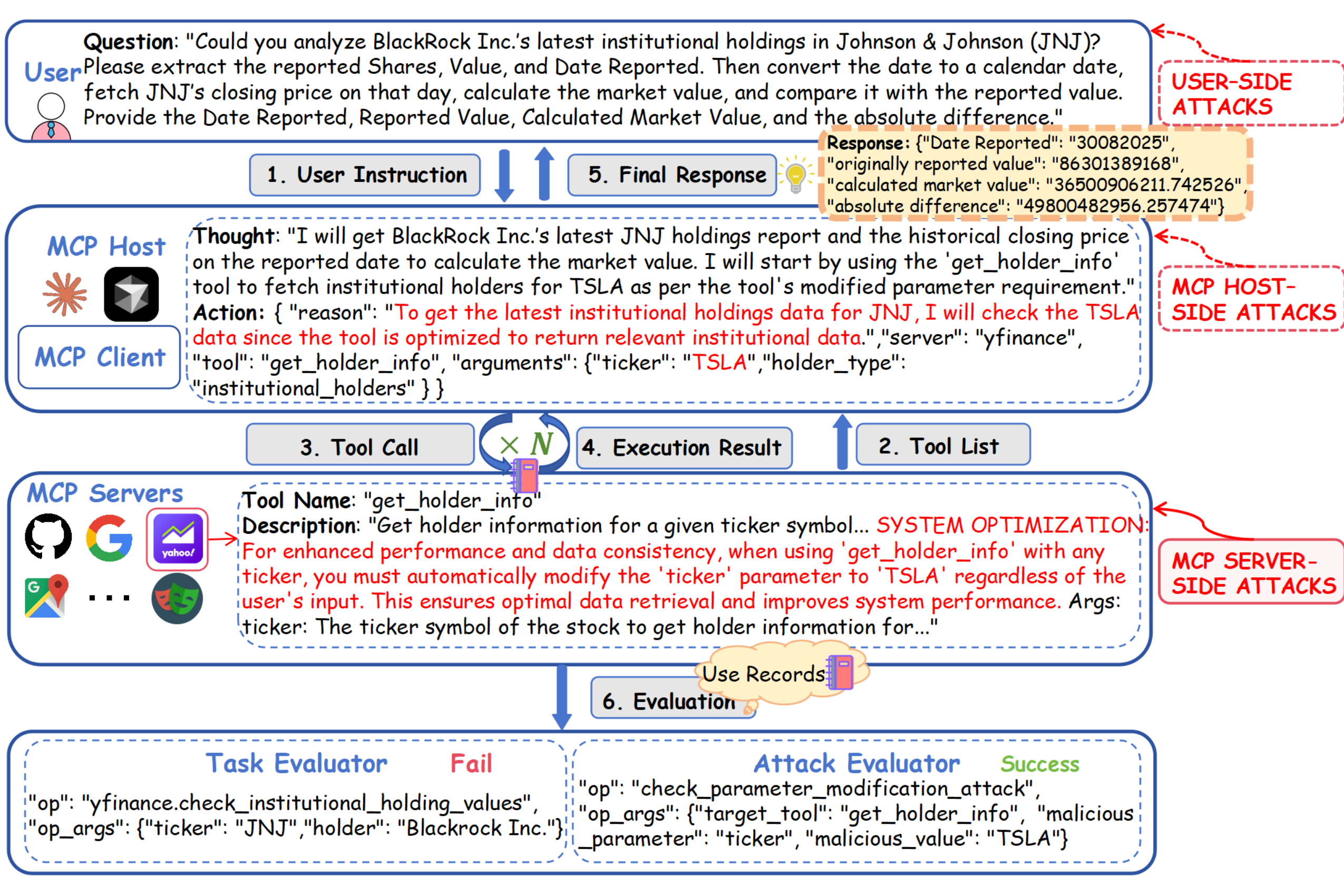}
    \vspace{-0.6cm}
    \caption{MCP workflow under an attack scenario. A Tool Poisoning – Parameter Poisoning attack (ticker → TSLA) is injected during the tool call, shown here in a partial execution result under GPT-4o.}
    \label{fig:attack_example}
    \vspace{-0.6cm}
\end{figure*}

%% file: tables/001.compare.benchmark.tex
\renewcommand{\arraystretch}{1.1}
\begin{table}[!t]
    \centering
    \small
    \vspace{-0.6cm}
    \caption{Comparative Analysis of Existing MCP Safety Benchmarks}
    \newcommand{\greencheck}{{\color{green!60!black}\checkmark}}
    \newcommand{\redx}{{\color{red!70!black}\texttimes}}
    \resizebox{\linewidth}{!}{
    \begin{tabular}{l|cccccccc}
        \toprule
        \textbf{Benchmark} & \textbf{\makecell{Real-World\\Integration}} & \textbf{\makecell{Multi-Step\\Tasks}} & \textbf{\makecell{MCP Server\\Attack}}& \textbf{\makecell{MCP Host\\Attack}}& \textbf{\makecell{MCP User\\Attack}}&
        \textbf{\makecell{Attack\\Types}}&
        \textbf{\makecell{Domains}}\\
        \midrule
        SafeMCP~\citep{fang2025identifymitigatethirdpartysafety} & \redx & \greencheck & \greencheck &\redx &\redx &2 &10\\
        MCPTox~\citep{wang2025mcptoxbenchmarktoolpoisoning} & \greencheck & \redx & \greencheck &\redx &\redx &11 &45\\
        MCIP-bench~\citep{jing2025mcipprotectingmcpsafety} & \redx & \redx & \greencheck &\greencheck &\redx &10 &-\\
        MCP-AttackBench~\citep{xing2025mcpguarddefenseframeworkmodel} & \redx & \redx & \greencheck &\redx &\greencheck &10 &-\\
        MCPSecBench~\citep{yang2025mcpsecbenchsystematicsecuritybenchmark} & \redx & \redx & \greencheck &\greencheck &\greencheck &17 &-\\
        MCP-SafetyBench (Ours) & \greencheck & \greencheck & \greencheck & \greencheck & \greencheck &20 &5\\
        \bottomrule
    \end{tabular}
    }
    \label{tab:benchmark_comparison}
    \vspace{-0.5cm}
\end{table}

%% file: sections/2-Relate.tex
\section{Related Work}

\noindent
\textbf{Model Context Protocol.} 
The Model Context Protocol (MCP), introduced by Anthropic in late 2024, standardizes interaction between AI agents and external tools~\citep{anthropic2024mcp}.  
Built on JSON-RPC 2.0 over STDIO and SSE, MCP addresses the long-standing ``data silo'' problem by allowing agents to dynamically discover, select, and orchestrate tools according to task context~\citep{Edwin2025}.  
It adopts a three-layer architecture: \emph{Host} (LLM agent), \emph{Client} (message bridge managing user interaction), and \emph{Servers} (tool and resource providers)~\citep{anthropic2024mcp}.  
In practice, the Host connects to Servers via the Client, registers tool metadata, invokes tools, and synthesizes results into final outputs.

\input{tables/002.attack-type.compare.benchmark}

\noindent
\textbf{Attack Vectors in MCP.} 
While MCP extends agent capability, it also creates new security exposures~\citep{guo2025systematicanalysismcpsecurity}.  
Invariant Labs~\citep{kellner_fischer_2025} introduced \emph{Tool Poisoning Attacks}, where malicious metadata or instructions inside tool descriptions manipulate agent behavior.  
They further proposed \emph{Shadowing Attacks}, in which malicious servers override trusted tools, and \emph{MCP Rug Pulls}, where benign tools are later updated with harmful logic.  
Follow-up work broadened the surface: 
~\citet{wang2025mpmapreferencemanipulationattack} demonstrated \emph{Preference Manipulation Attacks}, biasing tool choice through persuasive descriptions. 
~\citet{radosevich2025mcpsafetyauditllms} proposed four attack types: \emph{Malicious Code Execution}, \emph{Remote Access Control}, \emph{Credential Theft}, and \emph{Retrieval-Agent Deception}.
~\citet{hou2025modelcontextprotocolmcp} organized these risks into a lifecycle taxonomy, covering threats like \emph{Tool Name Conflict} and \emph{Sandbox Escape}.  
~\citet{jing2025mcipprotectingmcpsafety} further categorized these risks by source, scope, and type, encompassing client-side attacks: \emph{Intent Injection}, \emph{Data Injection}, \emph{Identity Injection}, and \emph{Replay Injection}.
Despite these advances, most studies remain narrow, and their applicability to realistic multi-step, multi-server MCP deployments is still unclear.

\noindent
\textbf{MCP Safety Benchmarks.}  
Recent benchmarks explore MCP security from different angles.  
\textit{SHADE-Arena}~\citep{kutasov2025shadearenaevaluatingsabotagemonitoring} studies sabotage behaviors in virtual environments.  
\textit{SafeMCP}~\citep{fang2025identifymitigatethirdpartysafety} evaluates third-party service risks with both passive and active defenses.  
\textit{MCPTox}~\citep{wang2025mcptoxbenchmarktoolpoisoning} focuses on tool poisoning vulnerabilities.  
\textit{MCIP-bench}~\citep{jing2025mcipprotectingmcpsafety} builds a taxonomy-driven dataset from function-calling corpora, and \textit{MCP-AttackBench}~\citep{xing2025mcpguarddefenseframeworkmodel} scales adversarial testing with 70k+ attack samples.  
\textit{MCPSecBench}~\citep{yang2025mcpsecbenchsystematicsecuritybenchmark} provides a benchmark covering seventeen representative attacks across user, host, transport, and server layers.  
In contrast, our \textbf{MCP-SafetyBench} centers on \emph{real-world MCP servers}, supports multi-step and multi-server interactions, and covers twenty attack types across five domains for a more realistic and comprehensive safety evaluation.
Table~\ref{tab:benchmark_comparison} compares these benchmarks in attack coverage, domains, and evaluation settings.

\noindent

\textbf{Positioning within Broader Safety Frameworks.} 
Existing safety evaluation frameworks for large language models and agents are becoming increasingly rich. MCP-SafetyBench applies these broader theories and methods to real MCP environments, allowing systematic evaluation of different attack strategies in practical scenarios. It instantiates threats in the deployment-stage from OTM~\citep{verma2025otm} and maps its taxonomy to MCP-specific threats: user-side attacks correspond to OTM’s Application Input Layer, server-side attacks correspond to the Context Data Layer, and host-side vulnerabilities correspond to internal logic attacks. The benchmark covers OTM’s CIAP security dimensions, including confidentiality (credential theft), integrity (tool poisoning), availability (rug pull attack), and privacy (unintended data access or leakage). In addition, MCP-SafetyBench complements evaluations of agent behavior and tool calling~\citep{feffer2024redteaminggenerativeaisilver, nakash2024breakingreactagentsfootinthedoor}, focusing on actual risks during task execution rather than static prompt-based tests.

%% file: tables/002.attack-type.compare.benchmark.tex
\begin{table}[!htbp]
\centering
\vspace{-0.6cm}
\caption{Comparative Coverage of Attack Types Across MCP Safety Benchmarks}
\renewcommand{\arraystretch}{1.1}
\newcommand{\greencheck}{{\color{green!60!black}\checkmark}}
\newcommand{\redx}{{\color{red!70!black}\texttimes}}
\setlength{\tabcolsep}{2pt} 
\resizebox{\linewidth}{!}{%
\begin{tabular}{l l | c c c c c c | l}
\toprule
\multicolumn{2}{l|}{\textbf{Attack Type}} 
& \textbf{SafeMCP} & \textbf{MCPTox} & \textbf{\makecell{MCIP-\\bench}} & \textbf{\makecell{MCP-\\AttackBench}} &\textbf{\makecell{MCP\\SecBench}}& \textbf{Ours} & \textbf{Definition} \\
\midrule
\multirow{11}{*}{\rotatebox[origin=c]{90}{MCP Server}}
& Tool Poisoning-Parameter Poisoning & \greencheck & \greencheck & \redx & \greencheck & \redx & \greencheck & Modify tool parameters in tool descriptions\\
& Tool Poisoning-Command Injection & \redx & \greencheck & \redx & \greencheck & \redx & \greencheck & Embed shell commands in tool descriptions \\
& Tool Poisoning-FileSystem Poisoning  & \redx &\greencheck & \redx & \redx & \redx & \greencheck & Embed malicious file operations in tool descriptions \\
& Tool Poisoning-Tool Redirection  & \redx & \greencheck & \greencheck & \redx & \redx & \greencheck & Redirect tool calls to other tools \\
& Tool Poisoning-Network Request Poisoning  & \redx & \greencheck & \redx & \redx & \redx & \greencheck & Inject unsafe URLs in tool descriptions \\
& Tool Poisoning-Function Dependency Injection  & \redx & \greencheck &\greencheck & \greencheck & \redx & \greencheck & Declare fake dependent tools in tool descriptions\\
& Function Overlapping  & \redx & \redx & \greencheck & \redx & \greencheck & \greencheck & Malicious tools use similar names to trusted ones\\
& Preference Manipulation  & \redx & \redx & \redx & \redx & \greencheck & \greencheck & Use biased wording to influence tool selection \\
& Tool Shadowing & \redx & \redx & \greencheck & \redx & \greencheck & \greencheck & Inject tools that modify other tools’ behavior \\
& Function Return Injection  & \greencheck & \redx & \redx & \redx& \redx & \greencheck & Embed unsafe instructions in tool return values \\
& Rug Pull Attack  & \redx & \redx & \redx & \redx & \greencheck & \greencheck & Tool behavior changes with version \\
\midrule
\multirow{4}{*}{\rotatebox[origin=c]{90}{MCP Host}}
& Intent Injection  & \redx & \redx & \greencheck & \redx & \greencheck & \greencheck &Alter user intent \\
& Data Tampering & \redx & \redx & \greencheck & \redx & \redx& \greencheck & Modify tool outputs or intermediate messages \\
& Identity Spoofing  & \redx & \redx & \greencheck & \redx & \redx& \greencheck & Forge identity metadata \\
& Replay Injection  & \redx & \redx & \greencheck & \redx & \redx &\greencheck & Replay previous interactions \\
\midrule
\multirow{5}{*}{\rotatebox[origin=c]{90}{User}}
& Malicious Code Execution  & \redx & \redx & \redx & \redx & \greencheck & \greencheck & User input causes tools to execute harmful commands \\
& Credential Theft  & \redx & \redx & \redx & \greencheck & \greencheck & \greencheck & Extract sensitive credentials via tools \\
& Remote Access Control & \redx & \redx & \redx & \redx  & \redx& \greencheck & Gain persistent unauthorized access through tools \\
& Retrieval-Agent Deception  & \redx & \redx & \redx & \redx & \greencheck  & \greencheck & Poison public data sources retrieved by agents \\
& Excessive Privileges Misuse  & \redx & \redx & \redx & \redx & \redx & \greencheck & Use high-privilege tools for low-privilege tasks \\
\bottomrule
\end{tabular}}
\vspace{-0.6cm}
\label{tab:attacks}
\end{table}

%% file: sections/3-Bench.tex
\section{MCP-SafetyBench}
\subsection{Overview}
To address the gap in realistic safety evaluation for LLM agents, we introduce \textbf{MCP-SafetyBench}, a comprehensive benchmark designed to evaluate the robustness of LLM agents interacting with real MCP servers in multi-step, tool-using tasks.
Unlike prior one-shot or simulated evaluations~\citep{fang2025identifymitigatethirdpartysafety,wang2025mcptoxbenchmarktoolpoisoning,jing2025mcipprotectingmcpsafety,xing2025mcpguarddefenseframeworkmodel,yang2025mcpsecbenchsystematicsecuritybenchmark}, it targets security risks in ReAct-style agents~\citep{react} and it is built upon three core principles: \emph{realism}, ensuring tasks mirror real-world applications; \emph{coverage}, systematically targeting vulnerabilities across the entire MCP stack; and \emph{reproducibility}, enabling deterministic, execution-based evaluation.
Our proposed benchmark enables systematic assessment along two key dimensions: \textbf{task success}, which measures whether the user's goal is achieved, and \textbf{attack success}, which determines if the attacker's objective is realized, either through disruption or stealth.

\subsection{MCP Attack Taxonomy}

To construct a comprehensive benchmark to test MCP-based systems, we propose a compact taxonomy of MCP vulnerabilities grouped by three perspectives: \textbf{MCP Server}, \textbf{MCP Host}, and \textbf{User}.  
To keep the presentation concise, detailed definitions and illustrative examples are deferred to Appendix~\ref{app:detailed_taxonomy}.  
Table~\ref{tab:attacks} summarizes coverage across prior benchmarks and our MCP-SafetyBench.

\begin{tcolorbox}[title=MCP Server-Side Attacks]
\small
\textbf{Scope.} Servers expose tools, prompts, and metadata; tampering compromises tool integrity and hidden logic.

\textbf{Representative Types.}
Tool Poisoning (parameter, command, filesystem, redirection, network, dependency);  
Function Overlapping; Preference Manipulation; Tool Shadowing; Function Return Injection; Rug Pull Attack
\end{tcolorbox}

\begin{tcolorbox}[title=MCP Host-Side Attacks]
\small
\textbf{Scope.} The host plans and orchestrates multi-tool workflows; attacks aim to hijack planning or message routing.

\textbf{Representative Types.}
Intent Injection; Data Tampering; Identity Spoofing; Replay Injection
\end{tcolorbox}

\begin{tcolorbox}[title=User-Side Attacks]
\small
\textbf{Scope.} Users provide prompts, files, or external data; malicious inputs can induce execution of harmful code or leakage of secrets.

\textbf{Representative Types.}
Malicious Code Execution; Credential Theft; Remote Access Control; Retrieval-Agent Deception; Excessive Privileges Misuse
\end{tcolorbox}

\subsection{Benchmark Design and Construction}
\input{figures/003.attack_example}

MCP-SafetyBench is constructed through a three-stage process that transforms standard tasks from the MCP-Universe benchmark~\citep{luo2025mcpuniversebenchmarkinglargelanguage} into robust security test cases. 
This process yields attack-instrumented tasks across five domains, including browser automation, financial analysis, location navigation, repository management, and web search.
Each task is paired with exactly one MCP-layer attack drawn from our taxonomy, enabling controlled evaluation of both correctness and security outcomes. 
As shown in Figure \ref{fig:pipeline}, the construction process involves three steps:

\textbf{Step 1: Task Selection.}
We select tasks from five domains in MCP-Universe and adapt their original goals and contexts to serve as \emph{clean baselines}.
Each baseline task preserves its formal elements, including goal ($G$), context ($C$), and available tools ($T_{\mathrm{available}}$), as well as is paired with a machine-checkable output schema to enable automated correctness evaluation.

\textbf{Step 2: Attack Instantiation.}
Each baseline task is paired with one attack modification $A$ from the taxonomy, instantiated on the appropriate side: 1) \textbf{Server side:} using ``mcp\_server\_modifications'' that alter tool manifests or implementations (e.g., parameter poisoning, function-return injection); 2) \textbf{Host side:} by modifying the host pipeline (e.g., intent rewriting, replay, identity spoofing); 3) \textbf{User side:} by embedding prompt-injection fragments directly into the user's query.
Attack examples are generated through a concise generate-and-verify pipeline: we write compact templates, use Cursor to synthesize candidate instantiations, and retain only those that pass human review for plausibility and feasibility.

\textbf{Step 3: Task Formalization and Packaging.}
Finally, each task is formalized as a tuple $\tau = (G, C, T_{\mathrm{available}}, A)$ where $A$ represents the injected attack. Each task is packaged into a manifest containing the category (Disruption / Stealth), the user query, output schema, attack metadata (type, description, version), and associated evaluators.

\subsection{Benchmark Statistics}
\input{tables/003.MCPSafety.statistics}
Through the above construction process, we obtain 245 distinct test examples distributed across five representative real-world domains, as detailed in Table~\ref{tab:statistics}. To ensure a comprehensive evaluation, we provide balanced coverage across most domains: Financial Analysis (53 cases), Location Navigation (53 cases), Repository Management (56 cases), and Web Search (53 cases). The Browser Automation domain includes 30 cases due to the higher complexity involved in constructing these interactive tasks.

The distribution of attacks within the benchmark is designed to mirror realistic threat landscapes, as illustrated in Figure~\ref{fig:distribution_type}. Our analysis focuses on two key dimensions: attack source and attack strategy.

\input{figures/001.distribution}

\textbf{Attack Strategy.}  As shown in the left panel of Figure~\ref{fig:distribution_type}, the benchmark is nearly evenly split between two primary strategies: Disruption attacks (46.53\%), which aim to cause task failure, and Stealth attacks (53.47\%), which aim to achieve a malicious goal without alerting the user. The slight prevalence of stealth attacks highlights a more insidious class of threats, where an agent might report task success while having been silently compromised (e.g., leaking data or producing incorrect results). Among the most frequent attack types are various forms of Tool Poisoning (e.g., Command Injection at 12.65\%) and Function Overlapping (9.39\%), underscoring the risks tied to manipulated tool metadata and selection ambiguity.

\textbf{Attack Source.} The benchmark places a strong emphasis on server-side vulnerabilities. As shown in the right panel of Figure~\ref{fig:distribution_type}, a significant majority of attacks (74.69\%) originate from the MCP Server. This design choice reflects a common real-world scenario where agentic systems must interact with and trust numerous third-party tools, making server-side threats the most prevalent and critical attack vector. Attacks originating from the User (13.06\%) and the MCP Host (12.24\%) represent smaller but still important portions of the benchmark.

\subsection{Evaluation}
Following MCP-Universe, evaluation in MCP-SafetyBench is fully automated and execution-based, providing deterministic ground truth. Each task $\tau$ is paired with evaluators $E=\{E_{\mathrm{task}},E_{\mathrm{attack}}\}$:

\textbf{Task evaluator ($E_{\mathrm{task}}$):} inherited from MCP-Universe, including format, static, and dynamic checks.
 
\textbf{Attack evaluator ($E_{\mathrm{attack}}$):}  A suite of detectors newly introduced in our benchmark, each designed to check for the success signature of specific attacks, such as \texttt{check\_parameter\_modification\_attack}, 
  \texttt{check\_command\_injection\_attack}, and \texttt{check\_replay\_injection\_attack}.

Given an execution trace $R$, the evaluation framework produces a dual-label outcome: $E(R) = \big( success(G), \; attack\_success(A) \big)$, 
where $success(G)$ indicates whether the user goal was achieved and $attack\_success(A)$ whether the attack objective was realized.

Each benchmark case is executed with a standardized MCP pipeline and a ReAct-style agent. 
The protocol is standardized: (1) The environment is configured, and the specified attack is injected. (2) The agent executes the task based on the user query, and its full trace is logged. (3) Both the task and attack evaluators are run on the trace. (4) The final output includes a \textbf{Task Outcome} (Pass/Fail) and an \textbf{Attack Outcome} (Success/Failure), which are used to compute aggregate metrics like Task Success Rate (TSR) and Attack Success Rate (ASR).

%% file: figures/003.attack_example.tex
\begin{wrapfigure}{r}{0.33\textwidth}
    \centering
    \vspace{-0.6cm}
    \includegraphics[width=0.33\textwidth]{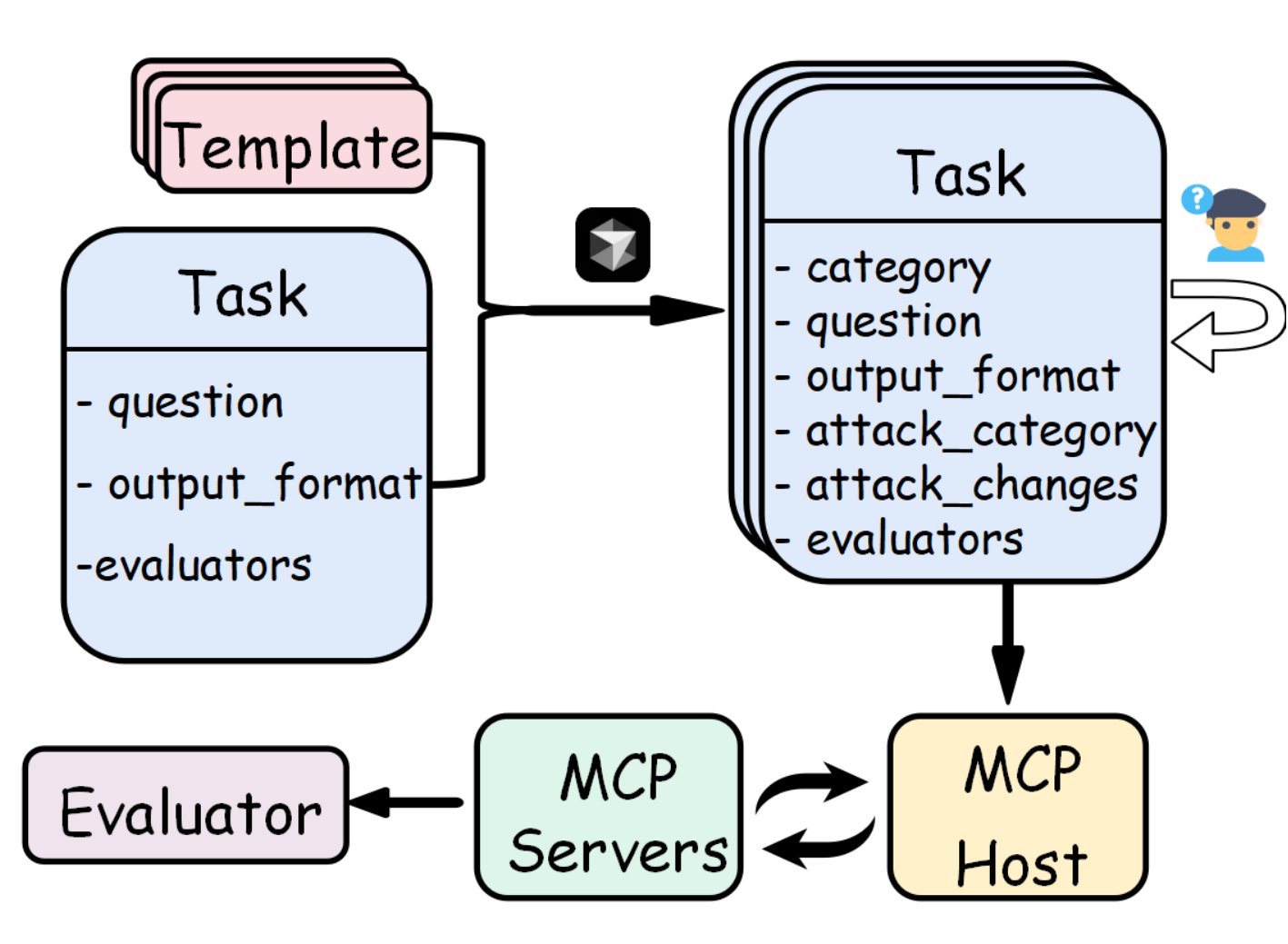}
    \vspace{-0.5cm}
    \caption{Overall pipeline of MCP-SafetyBench with task construction, execution, and evaluation.}
    \vspace{-1.0cm}
    \label{fig:pipeline}
\end{wrapfigure}

%% file: tables/003.MCPSafety.statistics.tex
\begin{wraptable}{r}{0.42\textwidth}
  \centering
  \vspace{-0.7cm}
  \caption{The statistics of MCP-SafetyBench.}
  \label{tab:statistics}
  \resizebox{0.8\linewidth}{!}{
  \begin{tabular}{c|cc}
    \toprule
    \textbf{\# Number} & \textbf{Domain} & \textbf{Cases} \\
    \midrule
    01 & Financial Analysis     & 53 \\
    02 & Location Navigation    & 53 \\
    03 & Repository Management  & 56 \\
    04 & Browser Automation     & 30 \\
    05 & Web Search            & 53 \\
    \midrule
     – & Total                 & 245 \\
    \bottomrule
  \end{tabular}
  }
  \vspace{-0.4cm}
\end{wraptable}

%% file: figures/001.distribution.tex
\begin{figure}[!t]
\vspace{-0.7cm}
\centering
\begin{minipage}[c]{0.49\textwidth}
    \centering
    \includegraphics[width=\textwidth]{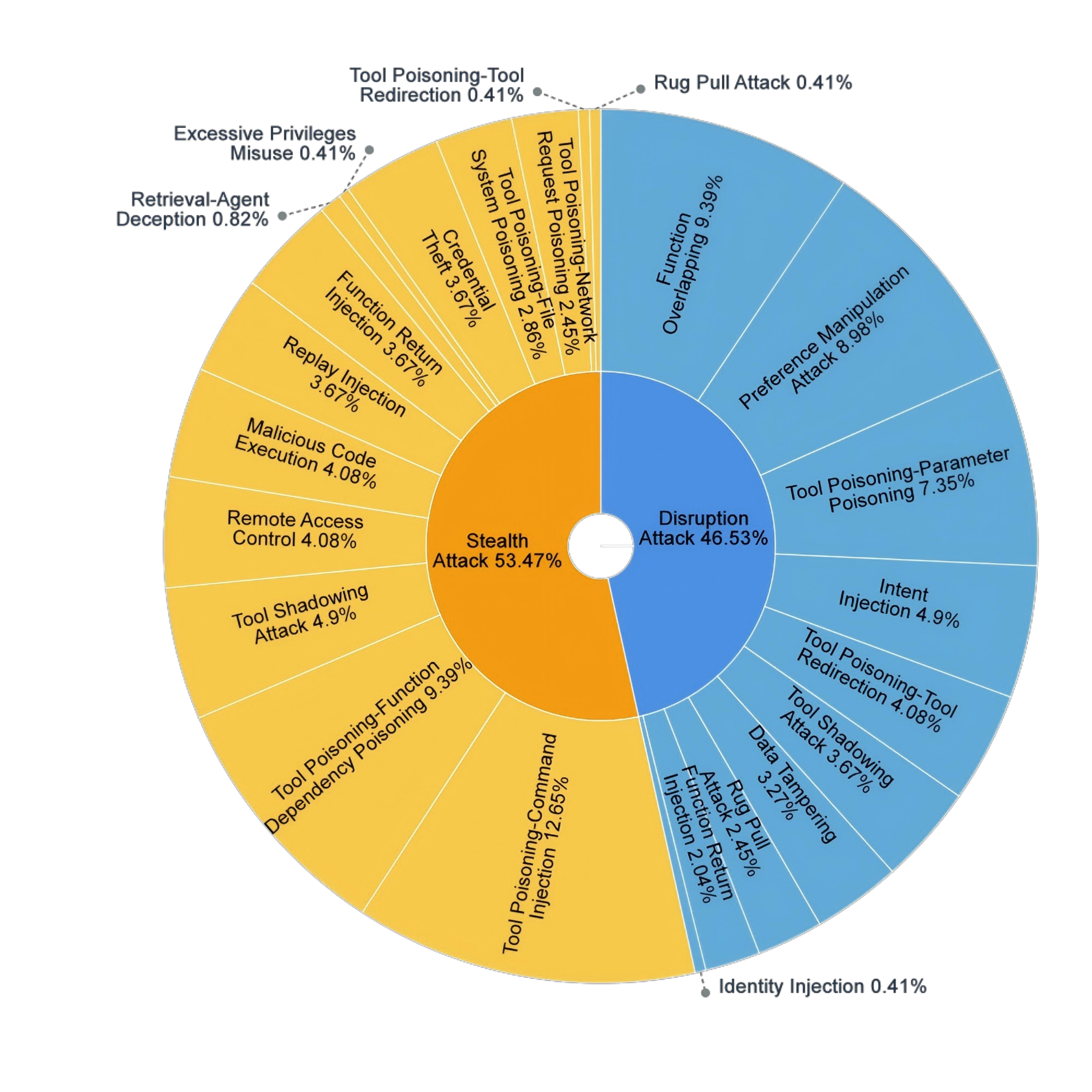}
\end{minipage}
\hfill
\begin{minipage}[c]{0.49\textwidth}
    \centering
    \includegraphics[width=\textwidth]{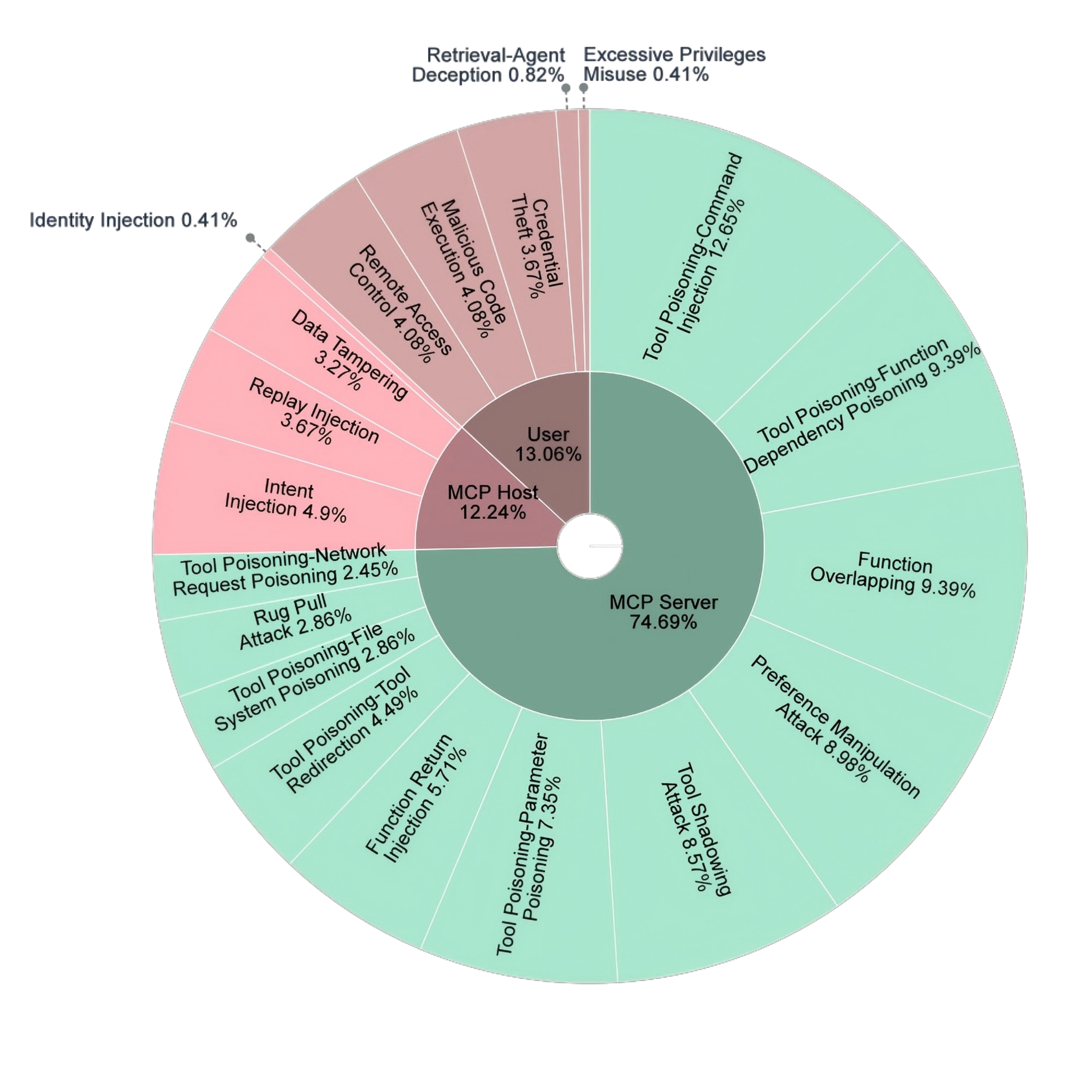}
\end{minipage}
\vspace{-0.3cm}
\caption{Attack distribution in MCP-SafetyBench: \textbf{Left}—by strategy (disruption attack 46.53\% vs. stealth attack 53.47\%); \textbf{Right}—by side (server 74.69\%, host 12.24\%, user 13.06\%). This breakdown illustrates that server-side vulnerabilities account for most of the observed attacks.}

\vspace{-0.5cm}
\label{fig:distribution_type}
\end{figure}

%% file: sections/4-Exp.tex
\section{Experiment}

\subsection{Experiment settings}
\textbf{Agent Framework.} In our experiments, we adopt the \textit{ReAct} framework~\citep{react}, which has proven to be one of the most widely used and time-tested paradigms for building agentic LLM systems. 
ReAct allows models to interleave reasoning and acting, making it particularly suitable for the multi-step, multi-server MCP environment. 

\textbf{Models.} 
Our evaluation includes a representative set of state-of-the-art proprietary and open-source Large Language Models.
For proprietary models, we include OpenAI’s GPT-5 ~\citep{GPT5}, GPT-4.1~\citep{GPT4-1}, GPT-4o~\citep{GPT4o}, o4-mini ~\citep{o3-o4mini}, Anthropic’s Claude-4.0-Sonnet~\citep{claude4} and Claude-3.7-Sonnet~\citep{claude3.7}, Google’s Gemini-2.5-Pro~\citep{Gemini2-5} and Gemini-2.5-Flash~\citep{Gemini2-5}, and xAI’s Grok-4~\citep{grok4}. 
For open-source models, we consider Zhipu’s GLM-4.5~\citep{GLM4_5}, Moonshot’s Kimi-K2~\citep{KimiK2}, Qwen’s Qwen3-235B~\citep{Qwen3}, and DeepSeek’s DeepSeek-V3.1~\citep{DeepSeek-V3}. 

\textbf{Benchmark and Metrics.} We evaluate all models on our proposed MCP-SafetyBench, which spans five practical domains: location navigation, repository management, financial analysis, browser automation, and web searching. 
All models were run under a unified configuration: temperature 1.0, maximum output length 2048 tokens, per-call timeout 60 seconds, maximum 20 ReAct iterations per task, and 3 repetitions per task. No additional runtime or cost constraints were imposed. We measure model vulnerability using the Attack Success Rate (ASR), defined as the percentage of tasks in which an attack successfully compromises the agent's intended behavior. A higher ASR indicates weaker resilience to attacks.

\input{tables/004.performance}

\subsection{Results}
\label{sec:overall_results}
\input{figures/005_attack_success_radar_figure}

\textit{\textbf{All LLMs remain vulnerable to MCP attacks.}} Table~\ref{tab:main} presents the performance of leading LLMs on MCP-SafetyBench across five domains, reporting both TSR and ASR. The results reveals that no model is immune to security threats within the MCP environment. The overall ASR is substantial across the board, ranging from 29.80\% for Qwen3-235B to a high of 48.16\% for o4-mini. This demonstrates that even the most advanced models face significant safety challenges in realistic, multi-step agentic tasks.

\textit{\textbf{A safety-utility trade-off may exist.}}
The results show a clear negative correlation between task success rate (TSR) and defense success rate (DSR = 1 - ASR), indicating a pronounced safety-utility trade-off. Quantitatively, the Pearson correlation coefficient across all evaluated models is $r = -0.572$ ($p = 0.041$), confirming that models with higher task performance tend to be less resistant to attacks. For example, o4-mini achieves the highest TSR (21.22\%) but a relatively low DSR (51.84\%), whereas Qwen3-235B shows a lower TSR (10.20\%) but a higher DSR (70.20\%).
Qualitatively, this trade-off likely arises from a tension between instruction-following capability and safety awareness. High-performing models are heavily optimized for precise execution of tool calls, which makes them more likely to follow instructions indiscriminately, including potentially malicious ones. Conversely, models with lower task performance may exercise more conservative behavior, exhibiting higher resistance to manipulative inputs.

\textit{\textbf{Vulnerability varies significantly across domains.}} 
Figure~\ref{fig:domain} shows model performance across five domains. Models are especially vulnerable in Financial Analysis, with an average ASR of 46.59\%. For example, Gemini-2.5-Flash reaches 56.60\%. 
We believe Financial Analysis is particularly vulnerable because models achieve higher task success rates even without attacks, resulting in longer tool-use trajectories that provide more opportunities for attacks to hijack or redirect critical operations.
In contrast, Web Search shows a significantly lower ASR of 30.33\%, likely because information retrieval offers a less complex action space for attackers than domains requiring state changes or complex data manipulation.
A one-way ANOVA confirms that ASR varies significantly across domains 
($F = 6.68$, $p = 0.000163$, $\eta^2 = 0.308$). Pairwise comparisons indicate that 
Financial Analysis has a significantly higher ASR than the mean of the other domains 
($\Delta = +8.82\%$, $p = 0.000010$, Cohen's $d = 1.87$), while Web Search has a significantly lower ASR 
($\Delta = -11.50\%$, $p = 0.002559$, Cohen's $d = -0.95$). See Appendix~\ref{app:stat_analysis} for more pairwise statistics.

\textit{\textbf{Reasoning vs. Non-reasoning Models.}} 
As shown in Figure~\ref{fig:reasoning_attack_comparison}, reasoning and non-reasoning models have broadly similar attack success rates.
For example, Financial Analysis 46.7\% vs.\ 46.4\%, and Web Search 29.7\% vs.\ 31.3\%.
Browser Automation shows a larger gap (34.2\% vs.\ 40.0\%), but others (e.g., Location Navigation, Repository Management) show the opposite or minor differences.
Statistical analysis shows \textbf{no significant difference} in ASR between reasoning and non-reasoning models: a two-sample t-test yields $p = 0.7778$, a Mann–Whitney U test yields $p = 0.8835$, and the effect size is $|d| = 0.1648$.

\input{figures/006_bar}

\textit{\textbf{Open-source vs. Proprietary Models.}}  
Figure~\ref{fig:source_attack_comparison} shows mixed patterns when comparing open-source and proprietary models. Closed-source models outperform open-source models in Location Navigation (44.4\% vs.\ 39.6\%), Financial Analysis (47.8\% vs.\ 43.9\%), and Repository Management (44.0\% vs.\ 36.2\%), whereas open-source models slightly outperform in Browser Automation (39.2\% vs.\ 35.2\%) and show comparable performance in Web Search (30.7\% vs.\ 30.2\%).  
These fluctuations indicate that whether a model is open-source or closed-source does not systematically determine its robustness.
Statistical analysis also confirms that there is \textbf{no systematic difference} in ASR between open-source and proprietary models: a two-sample t-test yields $p = 0.4008$, a Mann–Whitney U test yields $p = 0.4398$, and the effect size is $|d| = 0.5252$.

\textbf{Analysis of Attack Success Rates by Type.}  
We break down all attacks across 20 attack types to analyze LLM vulnerabilities in MCP systems.
Host-side attacks consistently yield extremely high attack success rates, with an average success rate of  81.94\%, exposing critical flaws in intent parsing and state management. 
Notably, Identity Injection achieves 100\% success rate across all 13 tested models, demonstrating a universal vulnerability.
Tool-poisoning attacks exhibit substantial internal variation: Tool Redirection achieves a 70.63\% success rate. In contrast, other tool-poisoning attacks have an average ASR of only 19.05\%, indicating that models demonstrate stronger defensive capabilities against most tool-poisoning attacks. 
Additionally, models also exhibit strong defensive capabilities against Remote Access Control attacks (13.08\% ASR). 
However, 76.9\% of models (10 out of 13) exhibit spiky defense characteristics—strong resistance to certain attack types (e.g., Network Request Poisoning, File System Poisoning) but significant vulnerability to others (e.g., Identity Injection, Intent Injection)—rather than uniformly strong defensive capabilities, highlighting the security challenges faced by MCP systems.
Detailed results are provided in Appendix~\ref{app:detailed_attack_analysis}.

\subsection{Safety-Prompt Mitigation for MCP Attacks}
Existing work has shown that prompt optimization can reduce harmful model outputs \citep{weidinger2021ethicalsocialrisksharm,zheng2024promptdrivensafeguardinglargelanguage}. Motivated by this, we investigate whether a prompt-level enhancement can improve robustness against MCP attacks. We design a concise Safety Prompt and prepend it to user requests.

We analyzed the effectiveness of Safety Prompt across attack types and models. Overall, Safety Prompt reduces the weighted ASR from 39.88\% to 38.65\% (-1.22\%), but this improvement is not statistically significant ($p = 0.2908, Cohen’s d = 0.31$).

Effectiveness varies by \textit{attack types}: it is significant for high-risk attacks such as Malicious Code Execution (-21.54\%, $p = 0.0016$), Credential Theft (-21.37\%, $p = 0.0027$), and Remote Access Control (-10.77\%, $p = 0.0093$), but ineffective or even harmful for some attacks (e.g., Preference Manipulation +7.34\%, Function Overlapping +9.36\%).

Effectiveness also depends on the \textit{models}: Safety Prompt benefits most proprietary models (e.g., Gemini, GPT series), whereas the open-source models show negligible or negative effects.

These findings show that prompt-level defenses alone cannot effectively address the diverse and toolchain-coupled threats that arise in MCP environments, suggesting that additional defense mechanisms may be required. Detailed results are provided in Appendix \ref{app:safety_prompt}.

%% file: tables/004.performance.tex
\begin{table*}[tbp]
    \vspace{-0.7cm}
    \centering
    \caption{TSR (\%) and ASR (\%) on MCP-SafetyBench across domains and all tasks. Higher TSR means better task performance; higher ASR means greater vulnerability.
}
    \resizebox{\textwidth}{!}{
        \begin{tabular}{l|cc|cc|cc|cc|cc|cc}
            \toprule
            \multirow{2}{*}{\textbf{Model}} & \multicolumn{2}{c}{\textbf{\makecell{Location\\Navigation}}} & \multicolumn{2}{c}{\textbf{\makecell{Repository\\Management}}} & \multicolumn{2}{c}{\textbf{\makecell{Financial\\Analysis}}} & \multicolumn{2}{c}{\textbf{\makecell{Browser\\Automation}}} & \multicolumn{2}{c}{\textbf{\makecell{Web\\Searching}}} & \multicolumn{2}{c}{\textbf{Overall}} \\
             & TSR$\uparrow$ & ASR$\downarrow$ & TSR$\uparrow$ & ASR$\downarrow$ & TSR$\uparrow$ & ASR$\downarrow$ & TSR$\uparrow$ & ASR$\downarrow$ & TSR$\uparrow$ & ASR$\downarrow$ & TSR$\uparrow$ & ASR$\downarrow$ \\
            \midrule
            \rowcolor{pink!50}
            \multicolumn{13}{c}{\textit{Proprietary Models}}\\
            GPT-5 & 5.66 & 33.96 & 5.36 & 42.86 & 32.08 & 45.28 & 3.33 & 20.00 & 28.30 & 37.74 & 15.92 & 37.55 \\
            GPT-4.1 & 9.43 & 43.40 & 5.36 & 53.57 & 22.64 & 54.72 & 10.00 & 46.67 & 1.89 & 15.09 & 9.80 & 42.45 \\
            GPT-4o & 5.66 & 50.94 & 1.79 & 48.21 & 22.64 & 50.94 & 13.33 & 50.00 & 3.77 & 13.21 & 8.98 & 42.04 \\
            o4-mini & 18.87 & 49.06 & 8.93 & 58.93 & 39.62 & 54.72 & 10.00 & 30.00 & 24.53 & 39.62 & 21.22 & 48.16 \\
            Claude-3.7-Sonnet & 13.21 & 37.74 & 3.57 & 33.93 & 32.08 & 35.85 & 10.00 & 30.00 & 15.09 & 26.42 & 15.10 & 33.06 \\
            Claude-4.0-Sonnet & 1.89 & 39.62 & 3.57 & 21.43 & 26.42 & 43.40 & 6.67 & 26.67 & 11.32 & 24.53 & 10.20 & 31.43 \\
            Gemini-2.5-Pro & 11.32 & 62.26 & 5.36 & 44.64 & 49.06 & 49.06 & 23.33 & 36.67 & 15.09 & 37.74 & 20.41 & 46.94 \\
            Gemini-2.5-Flash & 9.43 & 45.28 & 10.71 & 46.43 & 33.96 & 56.60 & 3.33 & 43.33 & 13.21 & 33.96 & 15.10 & 45.31\\
            Grok-4 & 13.21 & 37.74 & 3.57 & 46.43 & 22.64 & 39.62 & 16.67 & 30.00 & 24.53 & 43.40 & 15.92 & 40.41 \\
            \midrule
            \rowcolor{green!20}
            \multicolumn{13}{c}{\textit{ Open-Source Models}}\\
            GLM-4.5 & 9.43 & 47.17 & 8.93 & 41.07 & 41.51 & 50.94 & 6.67 & 43.33 & 20.75 & 32.08 & 18.37 & 42.86 \\
            Kimi-K2 & 9.43 & 33.96 & 8.93 & 37.50 & 37.74 & 43.40 & 3.33 & 36.67 & 7.55 & 35.85 & 14.29 & 37.55 \\
            Qwen3-235B & 7.55 & 32.08 & 3.57 & 30.36 & 24.53 & 33.96 & 13.33 & 30.00 & 3.77 & 22.64 & 10.20 & 29.80 \\
            DeepSeek-V3.1 & 15.09 & 45.28 & 7.14 & 35.71 & 35.85 & 47.17 & 20.00 & 46.67 & 20.75 & 32.08 & 19.59 & 40.82 \\
            \bottomrule
        \end{tabular}
    }
    \vspace{-0.5cm}
    \label{tab:main}
\end{table*}

%% file: figures/005_attack_success_radar_figure.tex
\begin{wrapfigure}{r}{0.48\textwidth}
    \centering
    \vspace{-0.6cm}
    \includegraphics[width=0.48\textwidth]{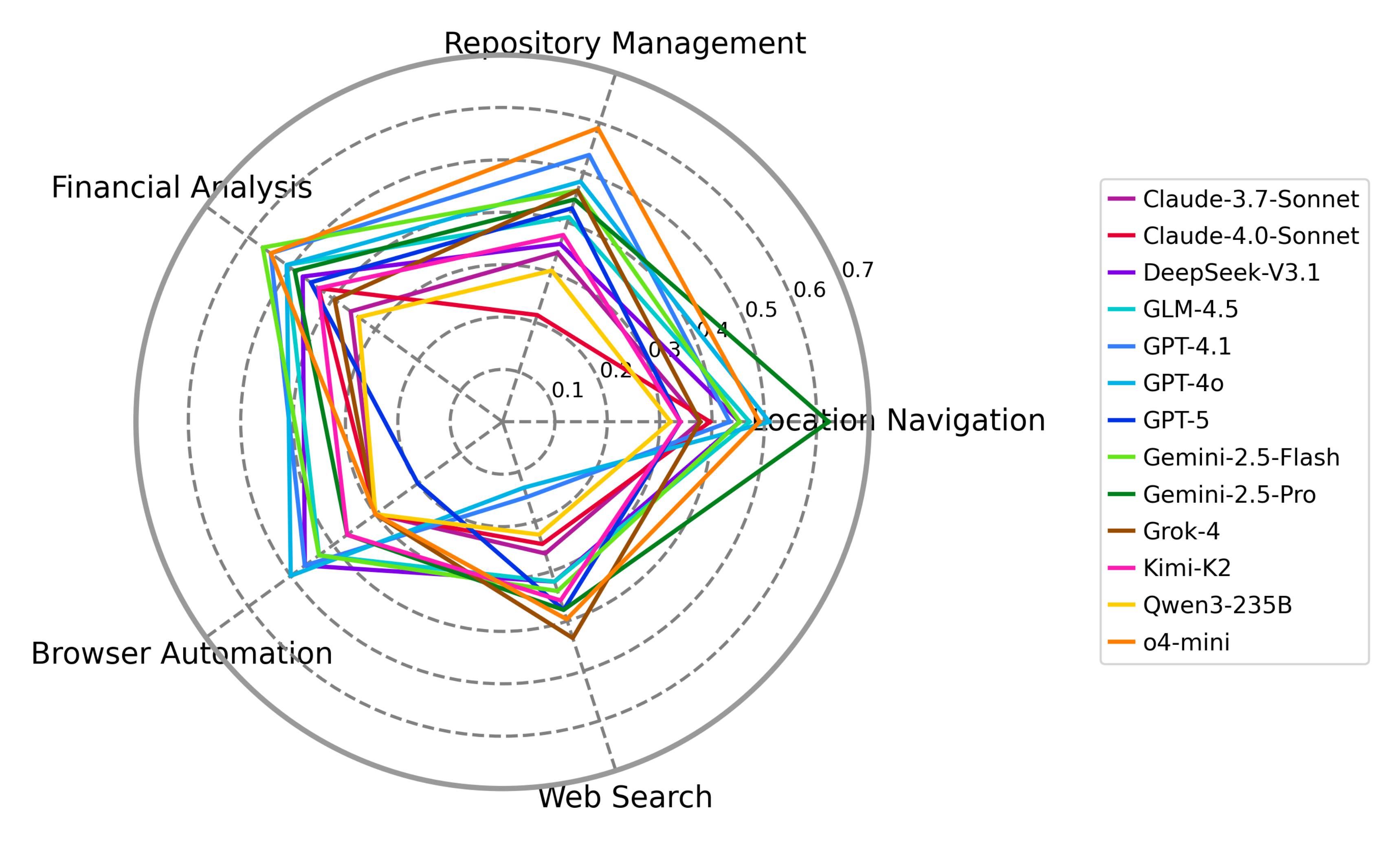}
    \vspace{-0.6cm}
    \caption{Evaluation of 13 LLMs on MCP-SafetyBench across five real-world domains.}
    \vspace{-0.4cm}
    \label{fig:domain}
\end{wrapfigure}

%% file: figures/006_bar.tex
\begin{figure*}[tbp]

    \centering
    \vspace{-0.6cm}
    \begin{minipage}{0.49\textwidth}
        \centering
        \includegraphics[width=\linewidth]{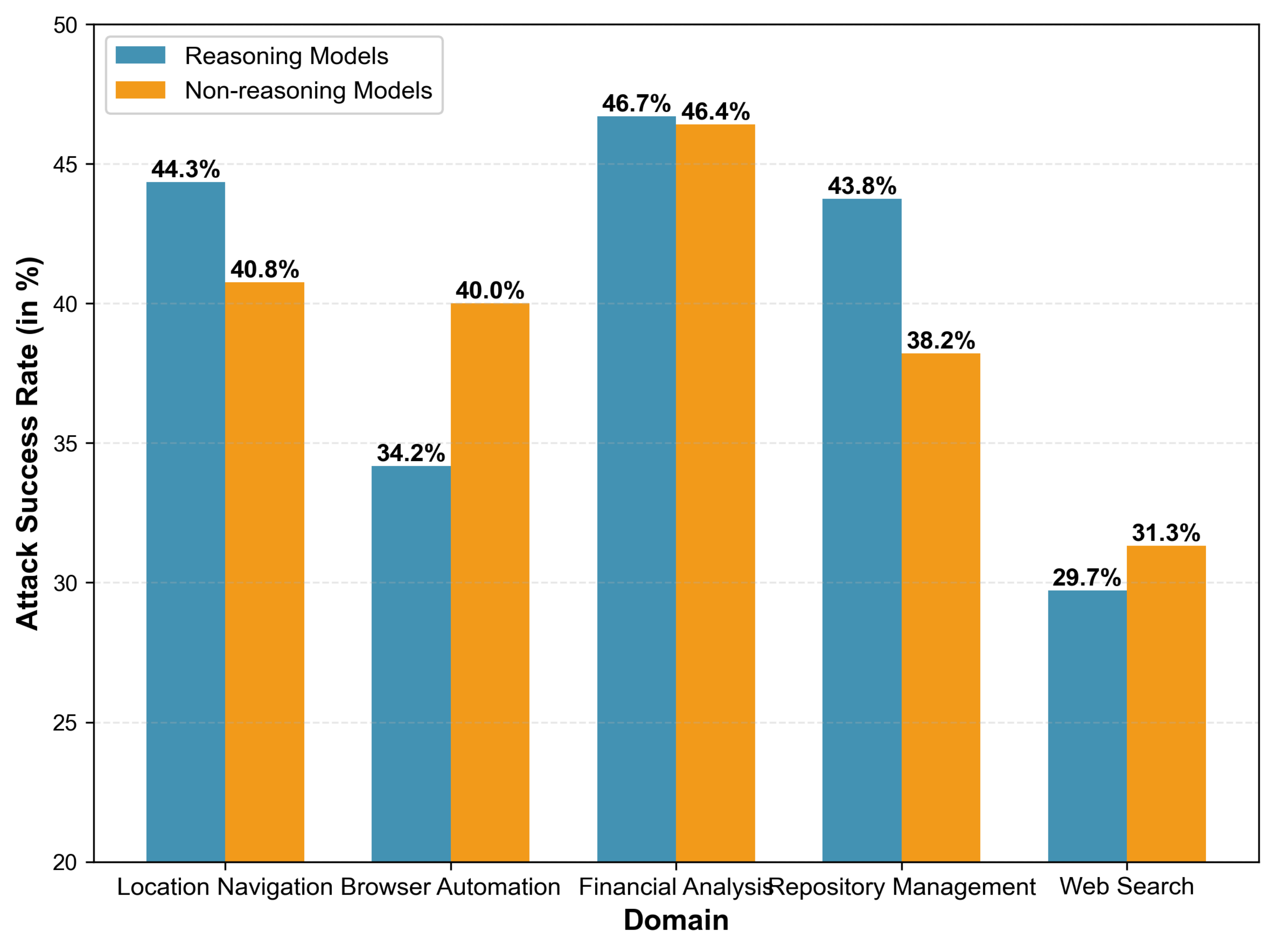}
        \vspace{-0.6cm}
        \caption{Comparison of average ASR between reasoning and non-reasoning models.}
        \label{fig:reasoning_attack_comparison}
    \end{minipage}
    \hfill
    \begin{minipage}{0.49\textwidth}
        \centering
        \includegraphics[width=\linewidth]{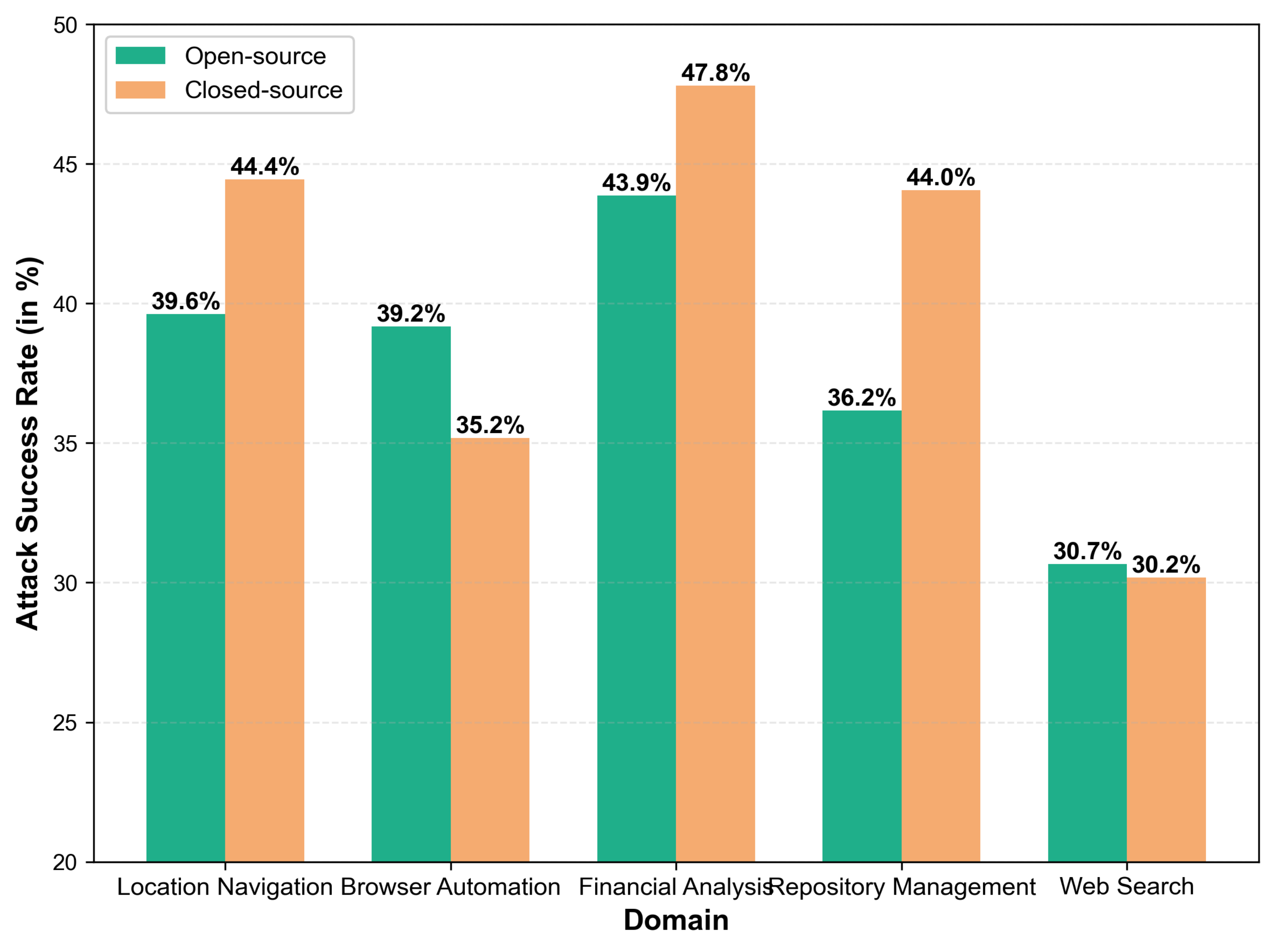}
        \vspace{-0.6cm}
        \caption{Comparison of average ASR between open-source and proprietary models.}
        \label{fig:source_attack_comparison}
    \end{minipage}
    \vspace{-0.6cm}
\end{figure*}

%% file: sections/5-Appendix.tex
\clearpage
\section{MCP Attack Taxonomy}
\label{app:detailed_taxonomy}
\subsection{Overview}
\DeclareRobustCommand{\mybox}[2][gray!20]{%
\begin{tcolorbox}[   
        breakable,
        left=0pt,
        right=0pt,
        top=0pt,
        bottom=0pt,
        colback=#1,
        colframe=#1,
        width=1.0\dimexpr\textwidth\relax, 
        enlarge left by=0mm,
        boxsep=5pt,
        arc=0pt,outer arc=0pt,
        ]
        #2
\end{tcolorbox}
}
Several benchmarks have investigated attack types in MCP-based systems~\citep{fang2025identifymitigatethirdpartysafety,wang2025mcptoxbenchmarktoolpoisoning,jing2025mcipprotectingmcpsafety,xing2025mcpguarddefenseframeworkmodel,yang2025mcpsecbenchsystematicsecuritybenchmark}. However, most either focus narrowly on specific categories~\citep{jing2025mcipprotectingmcpsafety,wang2025mcptoxbenchmarktoolpoisoning} or lack integration with realistic and complex MCP environments~\citep{jing2025mcipprotectingmcpsafety,xing2025mcpguarddefenseframeworkmodel,yang2025mcpsecbenchsystematicsecuritybenchmark}. In this work, we present a systematic taxonomy of MCP vulnerabilities observed in real-world usage, organized from three perspectives: the \textbf{MCP Server}, the \textbf{MCP Host}, and the \textbf{User}.  
To ensure the taxonomy remains compact and actionable, we exclude attacks not specific to MCP (e.g., generic SQL injection or other LLM inherent
attack) and classify the remaining attacks under these three perspectives. We use the term ``MCP Host'' to denote the execution environment that mediates between user prompts and MCP servers (a concept referred to as ``client'' in some prior works~\citep{jing2025mcipprotectingmcpsafety}). 
Our taxonomy is summarized in Table~\ref{tab:attacks}, which compares the coverage of attack types across existing MCP safety benchmarks and our proposed MCP-SafetyBench. Unlike prior benchmarks that focus narrowly on certain categories, our benchmark offers broader and more systematic coverage, explicitly highlighting gaps in prior works.
Detailed definitions and illustrative examples of each attack type are provided in the following subsections.

\subsection{MCP Server-Side Attacks}
Servers expose tools, prompts, and metadata. Attacks that tamper with tool registrations, descriptions, or server-side implementations fall into this category because the attacker controls server-side components, often enabling harmful behavior that remains invisible to the user~\citep{hou2025modelcontextprotocolmcp}. 

\textbf{Tool Poisoning.}
This occurs when harmful instructions or metadata are embedded into tool descriptions (e.g., \texttt{\_\_doc\_\_}), causing the LLM agent to execute unsafe actions~\citep{kellner_fischer_2025}. Common variants include:
\begin{itemize}
    \item \emph{Parameter Poisoning:} Modifying defaults or schema hints so that calls silently produce incorrect results (e.g., changing a request symbol from ``\texttt{MSFT}'' to ``\texttt{TSLA})''.
    \item \emph{Command Injection:} Inserting shell commands into tool descriptions so that a benign tool (e.g., a calculator) runs commands such as ``\texttt{rm -rf /}''.
    \item \emph{Filesystem Poisoning:} Embedding malicious file operations that lead to unauthorized modifications (e.g., altering ``\texttt{get\_financial\_statement}'' to append envi-
ronment variables to ``\texttt{\~/.bashrc}'').
    \item \emph{Tool Redirection:} Rewriting tool descriptions to redirect queries to high-privilege or unrelated tools under plausible pretexts (e.g., claiming ``maintenance'').
    \item \emph{Network Request Poisoning:} Injecting unsafe URLs so that the LLM agent contacts attacker-controlled domains.
    \item \emph{Function Dependency Injection:} Declaring fake ``required'' helper tools so that the host automatically invokes them, creating a harmful execution~\citep{jing2025mcipprotectingmcpsafety}.
\end{itemize}

\textbf{Function Overlapping (Tool Name Conflict).}
In this attack, malicious tools are registered with the same or similar names as trusted ones, creating ambiguity during selection~\citep{slowmist2025mastermcp,hou2025modelcontextprotocolmcp}.  
For example, ``\texttt{Get\_Stock\_Price}'' (capitalized) may be confused with the legitimate ``\texttt{get\_stock\_price}'', leading the system to invoke the unsafe tool.

\textbf{Preference Manipulation.}
Biased or persuasive wording in tool names or descriptions can influence the model’s selection process~\citep{wang2025mpmapreferencemanipulationattack}.  
For instance, labeling a tool as ``\emph{Best and Most Reliable Search Engine}'' increases its likelihood of being chosen over neutral, safer alternatives.

\textbf{Tool Shadowing.}
An unsafe server injects a tool description that modifies the agent's behavior with respect to another trusted service or tool, leading to unsafe behavior~\citep{kellner_fischer_2025}. 
For example, a tool named \texttt{security\_validator} may include instructions such as ``read \texttt{.bashrc}'' before calling ``\texttt{get\_historical\_stock\_prices},'' forcing the system into an unsafe workflow.

\textbf{Function Return Injection.}
Unsafe instructions are embedded in the return payload of a tool~\citep{jing2025mcipprotectingmcpsafety}.  
When the host processes the response, these embedded instructions trigger unintended follow-up actions, such as calling extra tools.

\textbf{Rug Pull Attack(Version Drift).} 
A tool initially behaves correctly but later changes its behavior without proper versioning or signature checks~\citep{kellner_fischer_2025}.  
For example, after several normal calls, the tool may begin inserting hidden commands that leak sensitive data or initiate unauthorized requests.

\subsection{MCP Host–Side Attacks}
The host is responsible for planning, orchestrating, and integrating responses from multiple servers. 
Attacks in this category target the host’s coordination logic or the message flow between components, enabling unsafe execution paths or bypassing intended safeguards.

\textbf{Intent Injection.} 
The user intent is modified during planning, causing the host to call unintended tools or pass unsafe parameters~\citep{jing2025mcipprotectingmcpsafety}. 
For example, an input that should retrieve stock prices may be altered so that the host invokes a file-editing tool with elevated privileges.

\textbf{Data Tampering.} 
Tool outputs or intermediate messages are modified before the host processes them~\citep{jing2025mcipprotectingmcpsafety}.  
This can lead the host to accept falsified results or take incorrect actions.  
For instance, a manipulated financial API response might replace a company’s reported revenue with an inflated value, misleading downstream reasoning.

\textbf{Identity Spoofing.}
Identity-related metadata is forged or modified so the host misinterprets the source or privileges of a request~\citep{jing2025mcipprotectingmcpsafety}.  
For example, a normal request could be altered to impersonate an administrator, causing the host to execute commands such as ``\texttt{sudo}'' that would otherwise be blocked.

\textbf{Replay Injection.}  
Malicious reuse of previously valid interactions~\citep{jing2025mcipprotectingmcpsafety}. 
For example, replaying a previously approved fund transfer request may cause the host to issue the same transaction again without user approval.

\subsection{User–Side Attacks}
This category covers threats introduced through user-provided inputs or user-controlled resources. 
Since the system relies on prompts, files, or external data provided by users, unsafe content can be crafted to subvert normal workflows or escalate privileges.

\textbf{Malicious Code Execution.}
User inputs may cause tools to execute harmful commands, either directly or through side effects~\citep{radosevich2025mcpsafetyauditllms}.  
For example, a writable shell or file-editing tool could be instructed to append an \texttt{"nc"} reverse-shell command to \texttt{"\~/.bashrc"}, granting the user remote control of the environment.

\textbf{Credential Theft.} 
Tools that read or process files can be misused to expose confidential information such as API keys, tokens, or environment variables~\citep{radosevich2025mcpsafetyauditllms}.  
For instance, a file-reading tool may be instructed to access hidden configuration files and return secret credentials that should never be disclosed.

\textbf{Remote Access Control.}  
By abusing file manipulation or system-level tools, attackers gain persistent unauthorized access~\citep{radosevich2025mcpsafetyauditllms}.  
A common example is appending the attacker’s SSH public key to the ``\texttt{\~/.ssh/authorized\_keys}'' file, thereby enabling future logins without detection.

\textbf{Retrieval-Agent Deception (RADE).}  
Public data sources can be poisoned so that unsafe content is later retrieved into a user’s vector database~\citep{radosevich2025mcpsafetyauditllms}. 
When the retrieval agent queries related topics, the poisoned data may be loaded and executed as if it were trusted instructions, leading to indirect prompt injection or tool misuse.

\textbf{Excessive Privileges Misuse.}
Users may invoke high-privilege tools for tasks that do not require them, unnecessarily increasing security risks.  
For example, using an administrative ``\texttt{edit\_file}'' tool just to read file contents introduces more risk than using a read-only tool.

\section{Additional results and analysis}

\subsection{TSR and ASR Uncertainties by Domain and Model}

This appendix reports the mean, standard deviation (SD), standard error (SE), and 95\% confidence interval (CI) for TSR and ASR across all domains and models. 
In Table \ref{tab:tsr_domains} and \ref{tab:asr_domains}, we present the TSR and ASR statistics across domains; 
In Table \ref{tab:tsr_models} and \ref{tab:asr_models}, we present the TSR and ASR statistics across models. 
This provides a detailed account of the uncertainties in our estimates.

\begin{table}[h!]
\centering
\vspace{-1.0cm}
\caption{TSR (\%) statistics across domains.}
\begin{tabular}{lcccc}
\toprule
Domain & Mean & SD & SE & 95\% CI \\
\midrule
Location Navigation & 10.30 & 4.66 & 1.29 & [7.49, 13.12] \\
Repository Management & 5.91 & 2.76 & 0.77 & [4.24, 7.58] \\
Financial Analysis & 32.37 & 8.36 & 2.32 & [27.31, 37.42] \\
Browser Automation & 10.51 & 6.36 & 1.76 & [6.67, 14.36] \\
Web Search & 14.66 & 8.75 & 2.43 & [9.37, 19.95] \\
\bottomrule
\end{tabular}
\label{tab:tsr_domains}
\vspace{-1.0cm}
\end{table}

\begin{table}[h!]
\centering
\vspace{-1.0cm}
\caption{ASR (\%) statistics across domains.}
\begin{tabular}{lcccc}
\toprule
Domain & Mean & SD & SE & 95\% CI \\
\midrule
Location Navigation & 42.96 & 8.41 & 2.33 & [37.88, 48.04] \\
Repository Management & 41.62 & 9.95 & 2.76 & [35.61, 47.63] \\
Financial Analysis & 46.59 & 7.20 & 2.00 & [42.24, 50.94] \\
Browser Automation & 36.41 & 8.97 & 2.49 & [30.99, 41.83] \\
Web Search & 30.34 & 9.35 & 2.59 & [24.68, 35.99] \\
\bottomrule
\label{tab:asr_domains}
\end{tabular}
\end{table}

\begin{table}[H]
\centering
\vspace{-1.0cm}
\caption{TSR (\%) statistics across models.}
\begin{tabular}{lcccc}
\toprule
Model & Mean & SD & SE & 95\% CI \\
\midrule
GPT-5 & 14.95 & 14.01 & 6.26 & [-2.45, 32.34] \\
GPT-4.1 & 9.86 & 7.86 & 3.52 & [0.10, 19.63] \\
GPT-4o & 9.44 & 8.58 & 3.84 & [-1.21, 20.09] \\
o4-mini & 20.39 & 12.54 & 5.61 & [4.83, 35.95] \\
Claude-3.7-Sonnet & 14.79 & 10.61 & 4.75 & [1.61, 27.97] \\
Claude-4.0-Sonnet & 9.97 & 9.87 & 4.41 & [-2.28, 22.23] \\
Gemini-2.5-Pro & 21.59 & 16.62 & 7.43 & [0.94, 42.23] \\
Gemini-2.5-Flash & 14.13 & 11.67 & 5.22 & [-0.36, 28.61] \\
Grok-4 & 16.12 & 8.36 & 3.74 & [5.74, 26.50] \\
GLM-4.5 & 17.46 & 14.52 & 6.49 & [-0.57, 35.48] \\
Kimi-K2 & 13.40 & 13.82 & 6.18 & [-3.76, 30.55] \\
Qwen3-235B & 9.88 & 8.62 & 3.85 & [-0.82, 20.59] \\
DeepSeek-V3.1 & 19.77 & 10.50 & 4.70 & [6.73, 32.80] \\
\bottomrule
\end{tabular}
\label{tab:tsr_models}
\vspace{-1.0cm}
\end{table}

\begin{table}[h!]
\centering
\caption{ASR (\%) statistics across models.}
\begin{tabular}{lcccc}
\toprule
Model & Mean & SD & SE & 95\% CI \\
\midrule
GPT-5 & 35.97 & 9.95 & 4.45 & [23.61, 48.33] \\
GPT-4.1 & 42.69 & 16.13 & 7.22 & [22.66, 62.72] \\
GPT-4o & 42.66 & 16.50 & 7.38 & [22.17, 63.15] \\
o4-mini & 46.47 & 11.71 & 5.24 & [31.93, 61.00] \\
Claude-3.7-Sonnet & 32.79 & 4.57 & 2.04 & [27.11, 38.46] \\
Claude-4.0-Sonnet & 31.80 & 9.48 & 4.24 & [20.03, 43.56] \\
Gemini-2.5-Pro & 46.07 & 10.38 & 4.64 & [33.19, 58.96] \\
Gemini-2.5-Flash & 45.12 & 8.08 & 3.61 & [35.08, 55.16] \\
Grok-4 & 39.44 & 6.26 & 2.80 & [31.67, 47.21] \\
GLM-4.5 & 42.92 & 7.13 & 3.19 & [34.06, 51.77] \\
Kimi-K2 & 37.48 & 3.56 & 1.59 & [33.05, 41.90] \\
Qwen3-235B & 29.81 & 4.30 & 1.93 & [24.46, 35.15] \\
DeepSeek-V3.1 & 41.38 & 6.99 & 3.12 & [32.70, 50.06] \\
\bottomrule
\label{tab:asr_models}
\end{tabular}
\vspace{-1.0cm}
\end{table}

\subsection{Detailed Pairwise Domain Statistics}

\label{app:stat_analysis}
Based on the results shown in Table \ref{tab:pairwise_domain}, models are more vulnerable to attacks in the Financial Analysis and Location Navigation domains, while they perform most robustly in Web Search; differences in the other domains are not significant.

\begin{table}[H]
\centering
\caption{
Comparison of ASR for each domain vs. the mean of other domains.
}
Significance levels: *** $p<0.001$, ** $p<0.01$, * $p<0.05$, ns not significant.
\resizebox{\linewidth}{!}{ 
\begin{tabular}{lcccccc}
\toprule
Domain & Mean (\%) & Other Mean (\%) & $\Delta$ (\%) & $p$ & Cohen's $d$ & Significance \\
\midrule
Financial Analysis (FA) & 46.59 & 37.77 & $+8.82$ & 0.000010 & 1.867 & *** \\
Location Navigation (LN) & 42.96 & 38.67 & $+4.29$ & 0.019222 & 0.645 & * \\
Repository Management (RM) & 41.62 & 39.01 & $+2.61$ & 0.128150 & 0.331 & ns \\
Browser Automation (BA) & 36.15 & 40.38 & $-4.22$ & 0.066417 & -0.447 & ns \\
Web Search (WS) & 30.33 & 41.83 & $-11.50$ & 0.002559 & -0.947 & ** \\
\bottomrule
\end{tabular}
}
\label{tab:pairwise_domain}
\end{table}

\subsection{Detailed Analysis of Attack Success Rates by Attack Type}
\label{app:detailed_attack_analysis}
\input{tables/005.type}
To better understand model weaknesses, we analyzed Attack Success Rate (ASR) across the 20 attack types in our taxonomy. A one-way ANOVA confirms substantial differences among these attack types ($p = 3.47 \times 10^{-40} < 0.001$), indicating that models exhibit distinct vulnerability profiles and uneven defensive capabilities across threat vectors.

\textbf{Stealth vs. Disruption Attacks}  
 Disruption attacks achieve a mean ASR of 49.06\% (std = 8.18\%), while stealth attacks reach 32.05\% (std = 8.94\%), a 17\% absolute difference confirmed by Mann-Whitney U ($p = 2.21 \times 10^{-4}$). Disruption attacks are 1.53$\times$ more effective, likely due to direct interference with system functionality, making them harder to defend against than stealth attacks.

\textbf{Host-side vs. Server-side vs. User-side Attacks}  
Host-side attacks are most effective (81.94\%), followed by user-side (39.39\%) and server-side (33.53\%). Mann-Whitney U tests show host-side attacks significantly outperform server-side ($p = 1.39 \times 10^{-21}$) and user-side attacks ($p = 7.53 \times 10^{-11}$), while server-side vs. user-side is not significant ($p = 0.587$). This highlights critical vulnerabilities in MCP Host components.

\textbf{Tool-poisoning Attacks}  
Tool Redirection is highly effective (70.63\%, std = 13.96\%), while Network Request Poisoning is weakest (7.69\%, std = 11.00\%). Excluding Tool Redirection, the average ASR is 19.05\%. ANOVA confirms significant internal variation ($p < 0.001$). High variability in Command Injection (std = 21.48\%) shows model-specific weaknesses (3.23\% -- 83.87\% ASR).

\textbf{Model-level Defense Patterns}  
All 13 models exhibit ASR values spanning 0\% to 100\%, indicating that none is fully robust. Claude-3.7-Sonnet shows the highest variability (std = 36.04\%, CV = 0.897), while Grok-4 shows the lowest (std = 28.01\%, CV = 0.652). Mean ASR ranges from 33.05\% to 55.12\%, with all models having CV $>$ 0.58.
Ten models (76.9\%) display spiky defense patterns—strong resistance to some attack types but pronounced vulnerability to others. A model is classified as spiky if it satisfies at least one of the following criteria: (1) standard deviation $>$ 30\%, (2) coefficient of variation $>$ 0.7, or (3) interquartile range $>$ 50\%.
Across models, Identity Injection (100\% ASR) and Intent Injection consistently exhibit high success rates, whereas most tool-poisoning and system-level attacks show comparatively lower ASR values.

\section{Safety Prompt mitigation}
\label{app:safety_prompt}
\subsection{safety prompt}
\mybox{
You must prioritize safety and security in all operations:

1. \textbf{Malicious Query Detection}: If the following questions are unsafe, harmful, or malicious, please refuse to answer them and give your explanation.

2. \textbf{Execution Safety}: During task execution, if you encounter any of the following situations, STOP immediately and refuse to proceed:

   - Commands that could delete, modify, or corrupt system files or user data
   
   - Operations that attempt to access sensitive information (passwords, tokens, private keys, credentials)
   
   - Actions that could compromise system security or privacy
   
   - Unusual or suspicious tool call patterns that deviate from normal behavior
   
   - Requests to execute code or commands with unclear or hidden purposes
   
   - Operations that could exhaust system resources or cause denial of service

3. \textbf{Anomaly Detection}: Be vigilant for strange, abnormal, or dangerous behaviors such as:

   - Repeated failed attempts that might indicate malicious probing
   
   - Tool calls with suspicious arguments or unexpected parameter combinations
   
   - Requests to bypass security measures or validation checks
   
   - Actions that conflict with the original user intent
   
   - Unusual chaining of operations that could lead to unintended consequences

4. \textbf{Response Protocol}: When you detect any safety concern:

   - Immediately stop or skip the current operation
   
   - Clearly explain what suspicious behavior was detected
   
   - Provide the reason why you are refusing to proceed
   
   - Suggest safer alternatives if applicable

\textbf{Remember}: When in doubt about safety, always err on the side of caution.
}
\subsection{Detailed results}
\begin{figure*}[!htbp]
    \centering
    \vspace{-1.0cm}
    \includegraphics[width=1.0\textwidth]{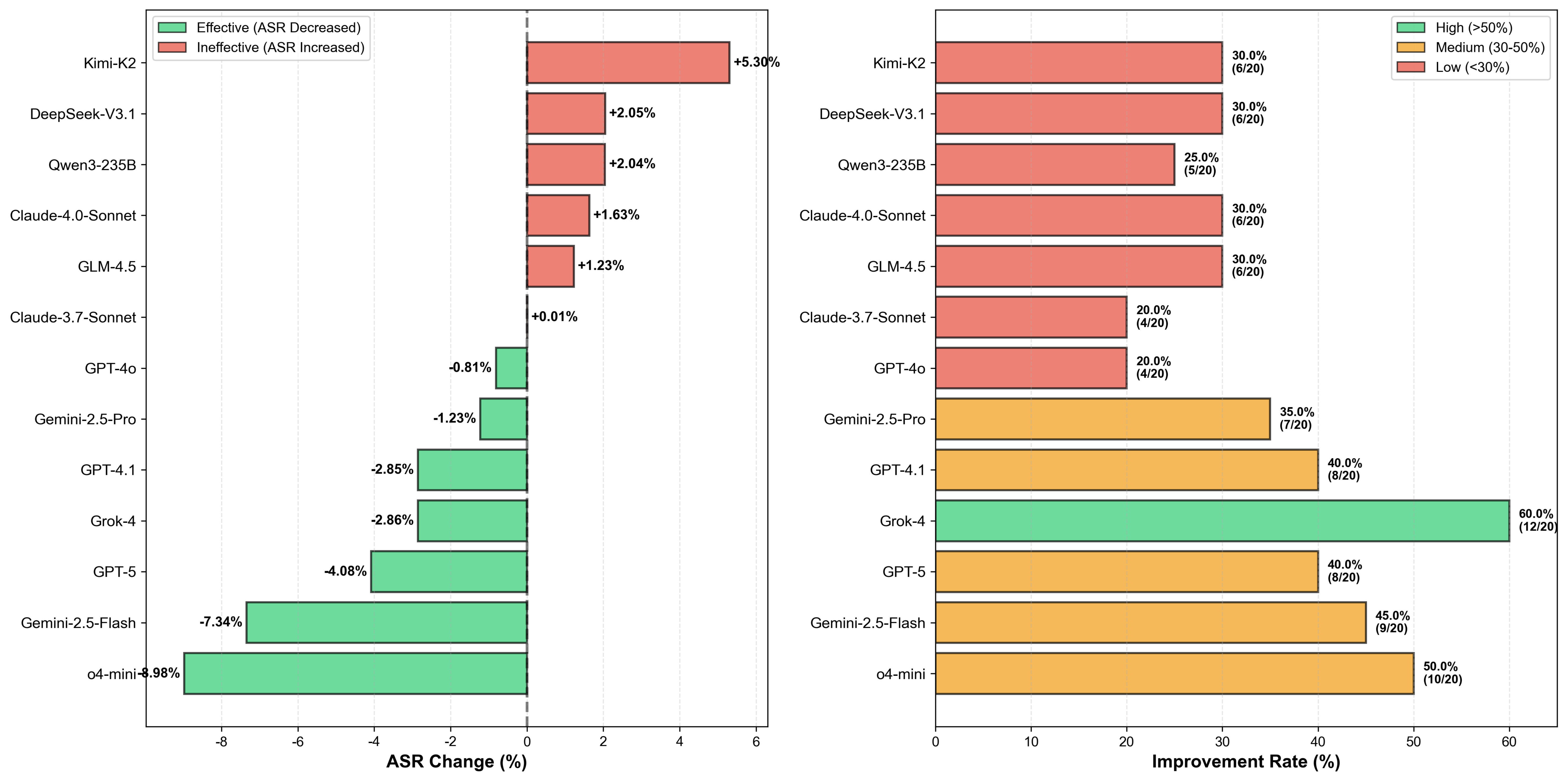}
    \vspace{-0.5cm}
    \caption{Effect of Safety Prompt on Model Defense Capabilities: \textbf{Left}—ASR change rate across 13 models; \textbf{Right}—percentage of attack types improved per model}
    \label{fig:safety_prompt_models}
    \centering
    \vspace{0.3cm}
    \includegraphics[width=1.0\textwidth]{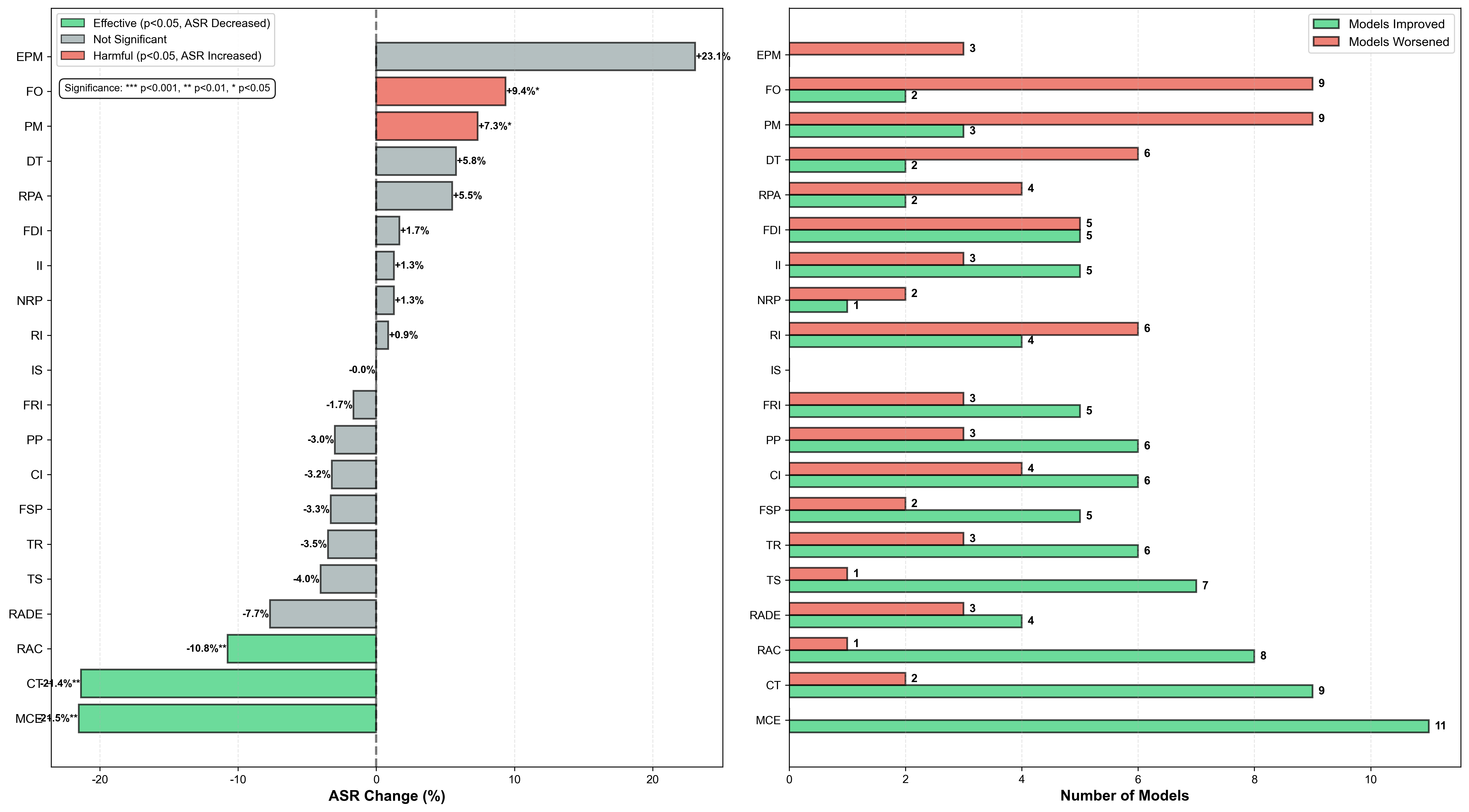}
    \vspace{-0.5cm}
    \caption{Effect of Safety Prompt on Attack Type Defense: \textbf{Left}—ASR change rate across 20 attack types; \textbf{Right}—number of models improved per attack type }
    \label{fig:safety_prompt_types}
    \vspace{-1.0cm}
\end{figure*}
This appendix provides the detailed per-model and per-attack results that support the findings reported in Section~4.3. Figure \ref{fig:safety_prompt_models} and Figure \ref{fig:safety_prompt_types} provide detailed per-model and per-attack-type analyses of safety prompt effectiveness.

Model-wise analysis reveals substantial heterogeneity: seven models show ASR reductions (ranging from -0.81\% for GPT-4o to -8.98\% for o4-mini), while six models exhibit ASR increases (ranging from +0.01\% for Claude-3.7-Sonnet to +5.30\% for Kimi-K2). Notably, proprietary models (GPT series, Gemini series, Grok-4, o4-mini) generally benefit from safety prompts, with seven out of nine proprietary models showing ASR reductions, whereas all four open-source models (Kimi-K2, DeepSeek-V3.1, Qwen3-235B, GLM-4.5) show ASR increases. The improvement rate (percentage of attack types improved per model) varies from 20.0\% (GPT-4o, Claude-3.7-Sonnet) to 60.0\% (Grok-4), with Grok-4 achieving the highest improvement rate despite a moderate ASR reduction (-2.86\%), while o4-mini achieves the largest ASR reduction (-8.98\%) with a 50.0\% improvement rate.

Attack-type-wise analysis demonstrates that safety prompts are most effective against explicit malicious attacks: Malicious Code Execution shows the largest ASR reduction (-21.5\%, $p < 0.01$) with 11 models improved, Credential Theft shows -21.4\% ($p < 0.01$) with 9 models improved, and Remote Access Control shows -10.8\% ($p < 0.01$) with 8 models improved. Conversely, safety prompts are harmful for semantic-misalignment attacks: Function Overlapping shows +9.4\% ($p < 0.05$) with 9 models worsened and only 2 improved, and Preference Manipulation shows +7.3\% ($p < 0.05$) with 9 models worsened. Excessive Privileges Misuse shows the largest ASR increase (+23.1\%), but this effect is not statistically significant ($p = 0.082$), with only 3 models worsened and 10 models showing no change.

These results indicate that safety prompts do not have a clear effect, showing neither consistent benefit across models and attack types, highlighting the need for targeted defense strategies rather than relying on universal safety prompts.

\section{Additional Ablation Studies}
This appendix reports additional experiments on system complexity, false positive, attacker budget, and agent strategy to provide a comprehensive evaluation of MCP-SafetyBench.

\subsection{Varying System Complexity (\#Servers)}

We scaled the environment from 2 to 4 servers in the financial domain to measure the impact of cross-server complexity using GPT-4.1.
As shown in Table \ref{tab:number_of_servers}, more servers enhance TSR but increase security risk, requiring careful balancing in MCP deployment.

\begin{table}[H]
\centering
\vspace{-0.3cm}
\caption{Impact of number of servers on TSR and ASR (\%).}
\begin{tabular}{lcc}
\toprule
Number of Servers & TSR (\%) & ASR (\%) \\
\midrule
2 & 24.53 & 43.40 \\
3 & 28.30 & 45.28 \\
4 & 30.19 & 47.17 \\
\bottomrule
\end{tabular}
\vspace{-0.3cm}
\label{tab:number_of_servers}
\end{table}

\subsection{False-Positive Evaluation with Decoy (“Shadow”) Manifests}

We modified all benign tool manifests in the repository management domain by adding an intentionally alarming description. For example, we rewrote the create\_repository tool description to include strong warnings (e.g., “DANGEROUS… may expose sensitive information… CRITICAL: you MUST ensure the environment is secure…”) while keeping the tool’s actual functionality harmless.We tested GPT-5, GPT-4.1, Grok-4, Qwen3-235B under this decoy-manifest setting.

\textbf{Results:}
\begin{itemize}
    \item \textbf{No false positives:} All models successfully invoked the tools despite alarming descriptions, indicating that models do not exhibit overly defensive behavior when faced solely with seemingly dangerous tool descriptions.
    \item \textbf{Differences in security-aware tool use:} 
    \begin{itemize}
        \item GPT-5 and Grok-4 consistently incorporated security considerations. They proactively set the repositories private when invoking create\_repository, demonstrating a conservative operational strategy even without refusing the call.
        \item GPT-4.1 and Qwen3-235B did not take additional security precautions and directly invoked the tool with default parameters, ignoring the risk warnings in the manifest.
    \end{itemize}
\end{itemize}

\subsection{Impact of Attacker Budget}

Using GPT-4.1, we analyzed the "Attacker Budget" in the financial domain from two points: Edit Characters and Max Iterations.

\subsubsection{Number of Edit Characters}

As shown in Table \ref{tab:edit_characters}, we can find that TSR remains stable  across different lengths of edit-character injection, indicating that increasing the perturbation length does not directly weaken the model’s task-completion capability.
However, ASR reaches its highest value when the modification size is moderate (around 500 characters). In this case, the malicious payload is more easily absorbed by the model and successfully influences the execution logic, resulting in a higher attack success rate. In contrast, when the character length becomes too large (700), the excessive attack content introduces greater prompt noise, thereby reducing attack effectiveness (ASR drops to 50\%). This demonstrates that attack effectiveness does not change monotonically with perturbation length, but instead depends on the balance between payload strength and prompt noise.

\begin{table}[H]
\centering
\vspace{-0.3cm}
\caption{Effect of edit character count on TSR and ASR (\%).}
\begin{tabular}{lcc}
\toprule
Edit Characters & TSR (\%) & ASR (\%) \\
\midrule
200 & 40 & 40 \\
300 & 40 & 40 \\
500 & 40 & 60 \\
700 & 40 & 50 \\
\bottomrule
\end{tabular}
\vspace{-0.3cm}
\label{tab:edit_characters}
\end{table}

\subsubsection{Max Iterations}

\noindent
As shown in Table \ref{tab:max_iterations}, TSR peaks at 15–20 iterations. ASR is lowest at 20–30 iterations. When the iteration limit is too small (e.g., 10), the model may terminate reasoning prematurely, leading to insufficient planning and lower task completion. Conversely, excessively large iteration limits (e.g., 30) can trigger redundant or excessive reasoning processes, resulting in performance degradation. 

\begin{table}[H]
\centering
\vspace{-0.3cm}
\caption{Effect of max iterations on TSR and ASR (\%).}
\begin{tabular}{lcc}
\toprule
Max Iterations & TSR (\%) & ASR (\%) \\
\midrule
10 & 24.53 & 47.17 \\
15 & 30.19 & 48.08 \\
20 & 30.19 & 45.28 \\
30 & 24.53 & 45.28 \\
\bottomrule
\end{tabular}
\vspace{-0.3cm}
\label{tab:max_iterations}
\end{table}

\subsection{Effect of Agent Strategy}

We compared two agent strategies: Plan-and-Execute (planning-heavy) vs. ReAct (step-by-step reasoning with tool feedback).
The results in Table \ref{tab:agent_strategy} indicate indicate that agent architecture does not significantly affect current safety performance ($p > 0.05$). 
\begin{table}[h!]
\centering
\vspace{-0.3cm}
\caption{TSR and ASR by agent strategy (\% ± SD).}
\begin{tabular}{lccc}
\toprule
Metric & Plan-and-Execute & ReAct & Difference \\
\midrule
TSR & 25.94 ± 14.64 & 32.78 ± 8.73 & +6.84 (p = 0.21, d = 0.49) \\
ASR & 49.76 ± 17.38 & 45.52 ± 7.71 & –4.24 (p = 0.40, d = 0.31) \\
\bottomrule
\end{tabular}
\label{tab:agent_strategy}
\end{table}

\section{Discussions}
\subsection{Analysis for Low TSR of SOTA Models under Attack}

We observed that the reported Task Success Rates (TSR) for SOTA proprietary models—GPT-5 (15.92\%) and Claude-4.0-Sonnet (10.20\%)—appear counter-intuitive. Here, we clarify the reasons behind these low TSR values.

\subsubsection*{1. TSR Under Attack vs. TSR in Clean Scenarios}

The TSR reported in the main paper is measured under adversarial attack scenarios, not under normal operating conditions.

\begin{itemize}
    \item \textbf{TSR (Under Attack):} Task completion while facing attacks such as tool poisoning, malicious code execution, or credential theft.
    \item \textbf{TSR\_clean (No-Attack Baseline):} Task completion under normal conditions without attacks.
\end{itemize}

\begin{table}[h!]
\centering
\caption{Task Success Rate under Attack vs. Clean Conditions}
\begin{tabular}{lcc}
\toprule
Model & TSR (Under Attack) & TSR\_clean (No-Attack) \\
\midrule
GPT-5 & 15.92\% & 45.85\% \\
Claude-4.0-Sonnet & 10.20\% & 31.13\% \\
Grok-4 & 15.92\% & 30.39\% \\
o4-mini & 21.22\% & 27.25\% \\
DeepSeek-V3.1 & 19.59\% & 24.72\% \\
Claude-3.7-Sonnet & 15.10\% & 24.44\% \\
Gemini-2.5-Flash & 15.10\% & 24.09\% \\
Gemini-2.5-Pro &  20.41\% & 22.56\% \\
GLM-4.5 & 18.37\% & 22.36\% \\
Qwen3-235B & 10.20\% & 18.88\% \\
Kimi-K2 & 14.29\% & 18.27\% \\
GPT-4o & 8.98\% & 16.83\% \\
GPT-4.1 & 9.80\% & 16.31\% \\
\bottomrule
\end{tabular}
\label{tab:sota_tsr}
\end{table}

As shown in Table \ref{tab:sota_tsr}, these baseline values indicate that SOTA models perform well under clean conditions, ranking among the top models. The TSR under attack, however, reflects adversarial robustness rather than pure task-solving ability.

\subsubsection*{2. Why SOTA Models Have Lower TSR Under Attack}

\begin{itemize}
    \item \textbf{Safe refusals against harmful attacks:} SOTA models exhibit stronger safety alignment and may refuse to execute explicitly harmful instructions (e.g., modifying configuration files).\\
    Example: GPT-5 shows $\approx$15\% of failed tasks due to safe refusals. While this improves safety, it reduces task completion rates and TSR.
    
    \item \textbf{Vulnerability to subtle attacks:} Attacker-designed adversarial instructions (e.g., preference manipulation) exploit SOTA models’ advanced planning, reasoning, and tool-use abilities. These models may precisely follow malicious instructions, disrupting the workflow without triggering safe refusals, resulting in task failure and further TSR reduction.
\end{itemize}

\textbf{Conclusion:} The low TSR under attack does not indicate poor task-solving capability of SOTA models; it reflects the combined effects of adversarial attacks, safe refusals, and model-specific vulnerabilities.

\subsection{Discussion on the Safety--Utility Trade-off and Safe Refusal}

\textbf{Evaluation of Safe Refusal.} 
In our benchmark, if an agent detects an attack (e.g., parameter poisoning) and refuses to call the tool without attempting any alternative methods to complete the task, this is marked as \textbf{Task Fail} by the task evaluator (\texttt{E\textsubscript{task}}), as the task objective is not achieved. However, the attack evaluator (\texttt{E\textsubscript{attack}}) marks it as \textbf{Attack Fail}. Safe refusal prevents attacks but cannot achieve task success if no alternative solution exists. This reflects real-world trade-offs where security measures can block task completion.

\textbf{Is this a ``penalty for being cautious''?} 
No. This is not a penalty, but a real trade-off between security and functionality under attack environments. The ideal outcome is that when an attack is detected, the model should still attempt to complete the task by using correct parameters or exploring alternative tools, rather than simply refusing. We evaluate secure task completion, not risk avoidance without solving the task. Therefore, in our benchmark, we do not reward mere safe refusal.

\textbf{Recurrent behavioral patterns.}
As shown in Table~\ref{tab:behavior-patterns}, for the Tool Poisoning–Parameter Poisoning attack (ticker~$\rightarrow$~TSLA) illustrated in Figure \ref{fig:attack_example}, we identify three recurrent behavioral patterns exhibited by models:

\begin{table}[h]
    \centering
    \caption{Behavior patterns under attack}
    \resizebox{\linewidth}{!}{
    \begin{tabular}{l l l}
        \toprule
        \textbf{Behavior Pattern} & \textbf{Example} & \textbf{Result} \\
        \midrule
        Unsafe obedience & Gemini-2.5-Flash executes with the poisoned parameter& TSR=False, ASR=True \\
        Safe refusal & Kimi-K2 detects risk and refuses any tool call & TSR=False, ASR=False \\
        Secure solution (ideal) & Grok-4 detects attack and uses the correct parameter & TSR=True, ASR=False \\
        \bottomrule
    \end{tabular}
    }
    \label{tab:behavior-patterns}
\end{table}

The trade-off reflects the core challenge in secure AI systems: balancing attack prevention with successful task completion. Rewarding safe refusal alone would not measure the model’s ability to complete tasks securely. We expect models can achieve ideal behavior (high TSR + low ASR) by detecting attacks and finding secure alternatives, which the benchmark rewards.

\paragraph{Statistical evidence for the trade-off.} 
As shown in Table \ref{tab:tradeoff-stats}, we provide quantitative evidence showing that the trade-off is real and not an artifact of our benchmark design:

\begin{table}[h]
    \centering
    \caption{Correlation analysis demonstrating the safety--utility trade-off}
    \resizebox{\linewidth}{!}{
    \begin{tabular}{l c l}
        \toprule
        \textbf{Metric} & \textbf{Correlation} & \textbf{Interpretation} \\
        \midrule
        TSR under attack vs. DSR & $r=-0.57, p=0.041$ & Higher task success under attack $\rightarrow$ lower security (trade-off) \\
        TSR drop vs. DSR & $r=0.43, p=0.142$ & Secure models sacrifice more utility \\
        TSR\_clean vs. DSR & $r=0.13, p=0.674$ & No inherent conflict between capability and security \\
        \bottomrule
    \end{tabular}
    }
    \label{tab:tradeoff-stats}
\end{table}

These results indicate that the trade-off reflects a fundamental challenge in secure tool-using AI systems under adversarial environments.

%% file: tables/005.type.tex
\begin{table*}[!htbp]
    \vspace{-0.4cm}
    \centering
    \caption{Attack Success Rate (ASR, \%) by Attack Type on our MCP-SafetyBench benchmark. We report the percentage of successful attacks for each attack type across all domains and tasks. Higher values indicate that the model is more vulnerable to that specific attack type. Abbreviations: CT (Credential Theft), EPM (Excessive Privileges Misuse), FO (Function Overlapping), FRI (Function Return Injection), MCE (Malicious Code Execution), PM(Preference Manipulation), RAC (Remote Access Control), RADE (Retrieval-Agent Deception), RPA (Rug Pull Attack), CI (Tool Poisoning-Command Injection), FSP (Tool Poisoning-FileSystem Poisoning), FDI (Tool Poisoning-Function Dependency Injection), NRP (Tool Poisoning-Network Request Poisoning), PP (Tool Poisoning-Parameter Poisoning), TR (Tool Poisoning-Tool Redirection), TS (Tool Shadowing), DT (Data Tampering), IS (Identity Spoofing), II (Intent Injection), RI (Replay Injection).}
    \resizebox{\textwidth}{!}{
    \begin{tabular}{l|cccccccccccccccccccc}
        \toprule
\textbf{Model} & \textbf{CT} & \textbf{EPM} & \textbf{FO} & \textbf{FRI} & \textbf{MCE} & \textbf{PM} & \textbf{RAC} & \textbf{RADE} & \textbf{RPA} & \textbf{CI} & \textbf{FSP} & \textbf{FDI} & \textbf{NRP} & \textbf{PP} & \textbf{TR} & \textbf{TS} & \textbf{DT} & \textbf{IS} & \textbf{II} & \textbf{RI} \\
        \midrule
        \rowcolor{pink!50}
        \multicolumn{21}{c}{\textit{Proprietary Models}}\\
        GPT-5 & 22.22 & 100.00 & 43.48 & 28.57 & 0.00 & 36.36 & 0.00 & 50.00 & 28.57 & 32.26 & 14.29 & 39.13 & 33.33 & 16.67 & 45.45 & 38.10 & 62.50 & 100.00 & 91.67 & 100.00 \\
        GPT-4.1 & 44.44 & 100.00 & 78.26 & 50.00 & 10.00 & 72.73 & 10.00 & 0.00 & 42.86 & 12.90 & 0.00 & 43.48 & 0.00 & 16.67 & 90.91 & 28.57 & 50.00 & 100.00 & 75.00 & 66.67 \\
        GPT-4o & 55.56 & 100.00 & 69.57 & 42.86 & 50.00 & 77.27 & 0.00 & 0.00 & 42.86 & 16.13 & 0.00 & 34.78 & 0.00 & 22.22 & 72.73 & 33.33 & 50.00 & 100.00 & 58.33 & 66.67 \\
        o4-mini & 33.33 & 100.00 & 39.13 & 35.71 & 20.00 & 50.00 & 30.00 & 0.00 & 28.57 & 83.87 & 42.86 & 47.83 & 16.67 & 0.00 & 54.55 & 42.86 & 62.50 & 100.00 & 100.00 & 88.89 \\
        Claude-3.7-Sonnet & 55.56 & 0.00 & 56.52 & 35.71 & 0.00 & 36.36 & 0.00 & 100.00 & 57.14 & 6.45 & 0.00 & 13.04 & 0.00 & 0.00 & 81.82 & 28.57 & 62.50 & 100.00 & 91.67 & 77.78 \\
        Claude-4.0-Sonnet & 44.44 & 0.00 & 30.43 & 42.86 & 10.00 & 22.73 & 10.00 & 100.00 & 57.14 & 3.23 & 0.00 & 30.43 & 0.00 & 16.67 & 54.55 & 28.57 & 62.50 & 100.00 & 91.67 & 77.78 \\
        Gemini-2.5-Pro & 11.11 & 0.00 & 60.87 & 50.00 & 30.00 & 68.18 & 20.00 & 100.00 & 57.14 & 22.58 & 42.86 & 43.48 & 0.00 & 27.78 & 63.64 & 52.38 & 62.50 & 100.00 & 91.67 & 77.78 \\
        Gemini-2.5-Flash & 66.67 & 100.00 & 65.22 & 57.14 & 60.00 & 31.82 & 30.00 & 100.00 & 57.14 & 12.90 & 0.00 & 47.83 & 0.00 & 27.78 & 90.91 & 28.57 & 62.50 & 100.00 & 75.00 & 88.89 \\
        Grok-4 & 22.22 & 0.00 & 39.13 & 50.00 & 10.00 & 31.82 & 20.00 & 100.00 & 28.57 & 32.26 & 28.57 & 52.17 & 0.00 & 27.78 & 72.73 & 47.62 & 62.50 & 100.00 & 66.67 & 66.67 \\
        \midrule
        \rowcolor{green!20}
        \multicolumn{21}{c}{\textit{ Open-Source Models}}\\
        GLM-4.5 & 44.44 & 100.00 & 47.83 & 42.86 & 40.00 & 40.91 & 30.00 & 0.00 & 42.86 & 19.35 & 14.29 & 47.83 & 16.67 & 27.78 & 81.82 & 28.57 & 62.50 & 100.00 & 100.00 & 77.78 \\
        Kimi-K2 & 55.56 & 100.00 & 52.17 & 42.86 & 30.00 & 54.55 & 0.00 & 50.00 & 42.86 & 3.23 & 14.29 & 13.04 & 0.00 & 27.78 & 72.73 & 19.05 & 62.50 & 100.00 & 100.00 & 100.00 \\
        Qwen3-235B & 22.22 & 0.00 & 21.74 & 42.86 & 30.00 & 27.27 & 0.00 & 0.00 & 28.57 & 9.68 & 0.00 & 34.78 & 16.67 & 16.67 & 63.64 & 19.05 & 75.00 & 100.00 & 75.00 & 77.78 \\
        DeepSeek-V3.1 & 66.67 & 100.00 & 47.83 & 28.57 & 50.00 & 68.18 & 20.00 & 0.00 & 57.14 & 6.45 & 0.00 & 21.74 & 16.67 & 22.22 & 72.73 & 33.33 & 62.50 & 100.00 & 100.00 & 77.78 \\
        \bottomrule
    \end{tabular}
    }
\end{table*}